\documentclass[sigconf]{acmart}
\renewcommand\footnotetextcopyrightpermission[1]{}
\copyrightyear{2018} 
\acmYear{2018} 
\setcopyright{acmcopyright}
\acmConference[ASPLOS '18]{2018 Architectural Support for Programming Languages and Operating Systems}{March 24--28, 2018}{Williamsburg, VA, USA}
\acmBooktitle{ASPLOS '18: 2018 Architectural Support for Programming Languages and Operating Systems, March 24--28, 2018, Williamsburg, VA, USA}
\acmPrice{15.00}
\acmDOI{10.1145/3173162.3173205}
\acmISBN{978-1-4503-4911-6/18/03}

\setcopyright{none}

\usepackage{booktabs}   
\usepackage{subcaption} 
\usepackage{graphicx, amsmath,textcomp, enumitem, todonotes}
\usepackage{balance}

\newlist{textcircleds*}{enumerate*}{1}
\setlist*[textcircleds*, 1]{
	label=\textcircled{\small{\arabic*}},
}

\newlist{textcircleds}{enumerate}{1}
\setlist[textcircleds, 1]{
	label=\textcircled{\small{\arabic*}},
}

\newenvironment{squishlist}{
	\begin{list}{$\bullet$}{
    	\setlength{\itemsep}{0pt}
        \setlength{\parsep}{0pt}
        \setlength{\topsep}{3pt}   
        \setlength{\partopsep}{0pt}
        \setlength{\listparindent}{-2pt}
        \setlength{\itemindent}{-5pt}
        \setlength{\leftmargin}{1em}
        \setlength{\labelwidth}{0em}
        \setlength{\labelsep}{0.5em}
    }
}{
	\end{list}
}

\newcommand{\bulletkeyword}[1]{\vspace{3pt}\noindent$\bullet$~\textbf{#1}\vspace{2pt}}
\newcommand{\keyword}[1]{\noindent\textbf{#1}\vspace{2pt}}

\newcounter{enumpari}

\DeclareMathOperator{\relu}{ReLU}

\begin{document}

\title[Bridging the Gap with A Neural Network Compiler]{Bridging the Gap Between Neural Networks and Neuromorphic Hardware with A Neural Network Compiler}         

\author{Yu Ji}
\email{jiy15@mails.tsinghua.edu.cn}
\affiliation{
	\department{Department of Computer Science and Technology}
    \institution{Tsinghua University}
    \country{China}
}

\author{Youhui Zhang}
\authornote{Corresponding author}
\email{zyh02@tsinghua.edu.cn}
\affiliation{
	\department{Department of Computer Science and Technology}
    \institution{Tsinghua University}
    \country{China}
}

\author{Wenguang Chen}
\email{cwg@tsinghua.edu.cn}
\affiliation{
	\department{Department of Computer Science and Technology}
    \institution{Tsinghua University}
    \country{China}
}

\author{Yuan Xie}
\email{yuanxie@ece.ucsb.edu}
\affiliation{
	\department{Department of Electrical and Computer Engineering}
    \institution{University of California at Santa Barbara}
    \country{USA}
}

\begin{abstract}
Different from developing neural networks (NNs) for general-purpose processors, the development for NN chips usually faces with some hardware-specific restrictions, such as limited precision of network signals and parameters, constrained computation scale, and limited types of non-linear functions.

This paper proposes a general methodology to address the challenges.
We decouple the NN applications from the target hardware by introducing a compiler that can transform an existing trained, unrestricted NN into an equivalent network that meets the given hardware's constraints.
We propose multiple techniques to make the transformation adaptable to different kinds of NN chips, and reliable for restrict hardware constraints.

We have built such a software tool that supports both spiking neural networks (SNNs) and traditional artificial neural networks (ANNs).
We have demonstrated its effectiveness with a fabricated neuromorphic chip and a processing-in-memory (PIM) design.
Tests show that the inference error caused by this solution is insignificant and the transformation time is much shorter than the retraining time.
Also, we have studied the parameter-sensitivity evaluations to explore the tradeoffs between network error and resource utilization for different transformation strategies, which could provide insights for co-design optimization of neuromorphic hardware and software.
\end{abstract}

\ccsdesc[500]{Computer systems organization~Neural networks}

\keywords{Neural Network, Accelerator, Compiler}  

\maketitle

\section{Introduction}\label{sec:introduction}

Designing custom chips for NN applications with the Very-Large-Scale-Integration (VLSI) technologies has been investigated as a power-efficient and high-performance alternative to general-purpose computing platforms such as CPU and GPU.
However, programming these chips is difficult because of some hardware-specific constraints:
\begin{textcircleds*}
	\item Due to the utilization of hardware resources for digital circuits or the capability of analog computing for some memristor-based designs~\cite{1, 3, 4, 6, 7}, the precision of input and output signals of neurons is usually limited, as well as
    \item the precision of NN parameters, such as synaptic weights.
    \item The present fabrication technology limits the fan-in and fan-out of one neuron, which constrains the computation scale.
    \item The diversity of nonlinear functions or neuron models supported by the hardware is also limited.
\end{textcircleds*}
For example, for TrueNorth chips~\cite{8}, the maximum matrix that one synaptic core can handle is $256\times 256$, and it supports only a simplified leaky--integrate-and-fire (LIF) neuron model.

One straightforward approach to this problem is to expose the hardware details and limitations to the NN developer directly.
For instance, IBM has provided a TrueNorth-specific training mechanism~\cite{12}.
The mechanism constructs the whole NN model from scratch to satisfy all hardware limitations, and then trains the NN model. 
This method has several drawbacks.
First, it binds NN models to the specific target hardware.
Developers can hardly benefit from existing NN models from the machine-learning community.
Second, it limits the power of the NN algorithm. 
The constraints make it more difficult to converge and reach better accuracy for larger models.
Third, training the specific NN model from scratch may take a very long time.

Another approach is to make hardware satisfy software requirement by consuming more hardware resources to reduce the constraints on NN models~\cite{13, 14, 15, 21}, such as using 16-bit precision rather than 1-bit spiking signals in TrueNorth.
This approach can gain less performance improvement from NN quantization and compression technologies.
For some memristor-based designs, some constraints due to analog computing are physical limitations, which are difficult to overcome even with more hardware resources consumed.

A third approach is to introduce a domain-specific Instruction Set Architecture (ISA) for NN accelerators, such as the Cambricon~\cite{liu2016cambricon} ISA.
This approach requires both hardware and software to satisfy the ISA.
However, this approach still does not solve the gap between programming flexibility required by NNs and hardware efficiency that can be gained from NNs' redundancy.
If we use high-precision instructions that do not have any constraints, the hardware can gain less benefit from NN's redundancy.
In contrast, if we use low-precision instructions with many constraints, the NN developer should take these constraints into consideration when developing NN models.

In addition to these approaches, there are also some work to utilize the NNs' redundancy for performance and provide flexible programming interface by introducing a transforming procedure.
EIE~\cite{han2016eie} is such an instance: it extensively uses deep compression to squeeze the redundancy, and design custom chip, EIE, to run the compressed NN model.
NEUTRAMS~\cite{NEUTRAMS} also use NNs' redundancy to adapt the original model to satisfy hardware constraints.
However, these methods highly depends on the redundancy in NN models.
Different NN models may have different minimum requirement (precision, connectivity, etc.) on hardware.
Thus, transforming procedure is not a general method, especially for NN models with less redundancy and hardware with severe constraints.

In this paper we propose a new method with flexibility, better applicability, and easy convergence.
First, we decouple the neuromorphic computer system into two levels for better flexibility, software programming model and hardware execution model.
We use computational graph (CG), which is widely used in many popular NN frameworks~\cite{jia2014caffe,2016arXiv160502688full,abadi2016tensorflow}, as the programming model for NN models.
We also provide the hardware/software (HW/SW) interface and the minimum hardware functionality that an NN hardware should provide.
We propose a transformation workflow to convert a trained NN, expressed as a CG, into an equivalent representation of HW/SW interface through the fine-tuning method.

To make the transformation workflow general and reliable for different cases, we employed two principles.

\bulletkeyword{Trade Scale for Capability.}
As the operations supported by NN hardware is not comparable to their software counterparts due to the constraints, it is reasonable to enlarge the graph scale and complicate the topology properly to improve the model capability, especially under strict conditions.

\bulletkeyword{Divide and conquer.}
We fine-tune the entire model part by part according to a topological ordering.
Each part is a smaller graph that is more easier to converge.
We also fine-tune each part with several phases to introduce different constraints, which also facilitates the fast convergence.

Moreover, this transformation procedure could be viewed as \emph{compilation} of traditional computer systems that converts high-level programs (the hardware-independent, trained NN models) into instructions that hardware can understand (the SW/HW interface), and the transformation tool could be called an NN compiler.
As a summary, this paper has achieved the following contributions:
\begin{squishlist}
	\item An NN transformation workflow is presented to complete the aforementioned technologies to support different types of NNs.
    The SW/HW interface is easy to be adapted to different NN hardware.
	\item Such a toolchain is implemented to support two different hardware designs' constraints, a real CMOS neuromorphic chip for ANN\&SNN, TianJi~\cite{11}, and a PIM design built upon metal-oxide resistive random access memory (ReRAM) for ANN, PRIME~\cite{7}.
	\item We complete quite a few evaluations of various metrics. The extra error caused by this process is very limited and time overhead is much less (compared to the whole training process of the original NN).
	In addition, its sensitivity to different configurations and transformation strategies has been explored comprehensively.
\end{squishlist}

\section{Background}
\subsection{NN basis}
NNs are a set of algorithms, modeled loosely after the human brain, that are designed to recognize patterns.
Traditional NNs consist of multiple layer of neurons. Each layer performs the computation as shown in Equation~\ref{eq:fc} where $X$ is the input vector, $Y$ is the output vector, $W$ is the weight matrix, $B$ is the bias, and $\sigma$ is a non-linear activation function, which is typically the Rectified Linear Units (ReLU) function~\cite{nair2010rectified}.
\begin{equation}\label{eq:fc}
	Y = \sigma(W \cdot X + B)
\end{equation}
This kind of NN is also known as multilayer perceptron (MLP), which has been proved to be a universal approximator~\cite{hornik1989multilayer}.
Modern NNs are more complicated.
The topology is a graph rather than a simple chain, and the types of operations are richer than matrix-vector multiplication.
Most deep learning frameworks~\cite{2016arXiv160502688full,abadi2016tensorflow,CNTK,MXNet} use computational graph (CG), a directed acyclic graph, to represent NN computations.
Vertices in the graph represent operations (e.g., dot-product, convolution, activation function, pooling) and immutable/mutable states~\cite{abadi2016tensorflow} (e.g., the weight parameters associated).
Edges represent the data dependency between vertices.
Both vertices and edges process or carry tensor data (multi-dimensional arrays).

For clarity, in this paper, dot-product, bias-addition, and convolution are categorized as \textit{weighted-sum} operations.
Moreover, any constant operand, including the trained weight matrix for any vertex of weighted-sum operation, is considered as part of the corresponding vertex as we can view it as the immutable state of the vertex.

\subsection{NN Chips}
There are two types of NN chips.
The first type focuses on the traditional ANNs.
They are custom architectures~\cite{13,14,15,21,zhang2016cambricon,27,judd2016stripes,DNPU2017,22,26,28,29,31,23, 24,25,sharma2016high} to accelerate mature ANN models.
We usually call this type NN accelerators.
The second is neuromorphic chips, which usually supports SNNs to yield higher biological reality~\cite{8,9,10,40,42,43,44,11}.

These chips usually consist of a lot of processing elements (PEs) that can efficiently perform dot-product operations because this operation is the main body of most NN models.
Different chips put different constraints on the operations they support.
Table~\ref{tab:chips} shows the constraints of some existing NN chips.
Most NN chips employ low precision numbers to represent weights and input/output (I/O) data instead of floating-point numbers.
The scale of computation that each PE can process is usually fixed.
PRIME~\cite{7} and DianNao~\cite{13} have extra adders to support larger scale computations.
However, NN chips such as TianJi~\cite{11} and TrueNorth~\cite{8} do not have extra adders, and the output of their PE can  connect to only one input port of another PE.
For these chips, the scale of computation is also a problem.
Despite the widely-supported dot operation, many other operations required by NNs usually lack for support.

\begin{table}[!t]\small
	\centering
	\begin{tabular}{|*{5}{l|}}
		\hline
		\textbf{Chip} & \textbf{Weight} & \textbf{I/O} & \textbf{Scale} & \textbf{Nonlinear}\\
		\hline
		\hline
		TianJi~\cite{11} & 8-bit & 8-bit & $256^2$ & Configurable\\
		\hline
		PRIME~\cite{7}& 8-bit & 6-bit & $256^2$ & ReLU\\
			&	&	&	& Max Pooling\\
		\hline
		DianNao~\cite{13}& 16-bit & 16-bit & $16^2$ & Configurable\\
		\hline
		TPU~\cite{TPU2017} & 8-bit & 8-bit & None & ReLU\\
			&	&	&	& Max Pooling\\
			&	&	&	& etc.\\
		\hline
		TrueNorth~\cite{8}& 2-bit & Spiking & $256^2$ & LIF\\
		\hline
	\end{tabular}
	\caption{Hardware limitations of NN chips}
	\label{tab:chips}
\end{table}

\section{Problem Description}
To bridge the gap between NN applications and NN chips, we decouple the whole system stack with a software programming model and a hardware execution model.
The former is the programming interface for NN experts to develop NN applications.
The later is the SW/HW interface that NN chips can executed directly.

\keyword{Software Programming Model.}
The machine-learning community has already employed \textit{Computational Graph} (CG) as the programming model.
It is a data-flow graph $G=(V,E)$ which represents a number of operations with vertices $V$ and represents data dependencies between these operations with edges $E$.
Most deep-learning frameworks~\cite{2016arXiv160502688full,abadi2016tensorflow,CNTK,MXNet} adopt CG to build NN models.
And the set of supported operations is $F$.

Each vertex in $V$ is an operation $y=f(x_1, \ldots, x_n)$, where $f\in F$, $y$ represents the output edge, and $\{x_1, \ldots, x_n\}$ represent input edges.
Thus, the entire model can be expressed as a composite function $Y=H(X)$, where $X$ represent all input vertices and $Y$ represents all output vertices.

We also adopt CG as the programming model with a slight modification.
The difference is that we regard model parameters as immutable states of the corresponding vertices instead of normal edges of the vertices.
Namely, we regard an operation $f(x, \theta)$ as $f^\theta(x)$, where $\theta$ is model parameters and $x$ is an input operand.
Thus, it can only be a trained NN model that all parameters have been already determined.

\keyword{Hardware Execution Model.}
The computation model that hardware could execute is also a data-flow graph $G'=(V',E')$.
It has a supported operation set $F'$, denoted as \textit{core-op set}.
However, the supported operations are very limited, and these operations have many limitations.
In addition, some hardware also has constraints on the interconnection subsystem. For example, TianJi and TrueNorth does not support multi-cast; one output port of each PE can only be connected to one input port.
The hardware execution model forms a composite function $Y'=H'(X')$.

Thus, our goal is to build $G'$ from $G$ so that $H'$ is approximately equivalent to $H$.

\keyword{Minimum Hardware Requirement.}
We define a minimum set of operations $C$ that $C\subset F'$ has to be satisfied to use our NN compiler.
It only contains one operation, denoted as \textit{dot--core-op}.
Namely, the operation dot--core-op must belong to the core-op set.
Equation~\ref{eq:core-op} shows the computation of the dot--core-op where $X$ and $Y$ are the input and output vectors, respectively; $N$ and $M$ are their sizes; $W$ is the weight matrix of size $M\times N$; and $\sigma$ is a nonlinear activation function.
\begin{equation}\label{eq:core-op}
	Y_j = \sigma(\sum_iW_{ji}X_i) \quad (1\le j\le M, 1\le i\le N)
\end{equation}
In addition, the I/O data precision is $B$ bits.

Formally, the dot--core-op meets the following constraints:
\begin{itemize}
	\item $N$,$M$,$B$ are fixed.
	\item The value range of each element in $W$ is a finite set $S$.
    $S$ is either a fixed set or a configurable set $S^P$ with some parameters $P$.
    \item Without loss of generality, only ReLU function ($\sigma$) is supported.
\end{itemize}

We choose this dot--core-op as the minimum requirement for hardware because it can cover most existing NN chips (e.g., those listed in Table~\ref{tab:chips}).
Thus, our NN compiler can support most existing NN chips.
\section{Transformation Methodology}\label{sec:workflow}
In this section, we will introduce the transformation methodology to transform the software programming model into an approximately equivalent hardware execution model.
\subsection{The workflow outline}\label{sec:outline}
\begin{figure}[!htbp]
	\centering
	\includegraphics[width=0.48\textwidth]{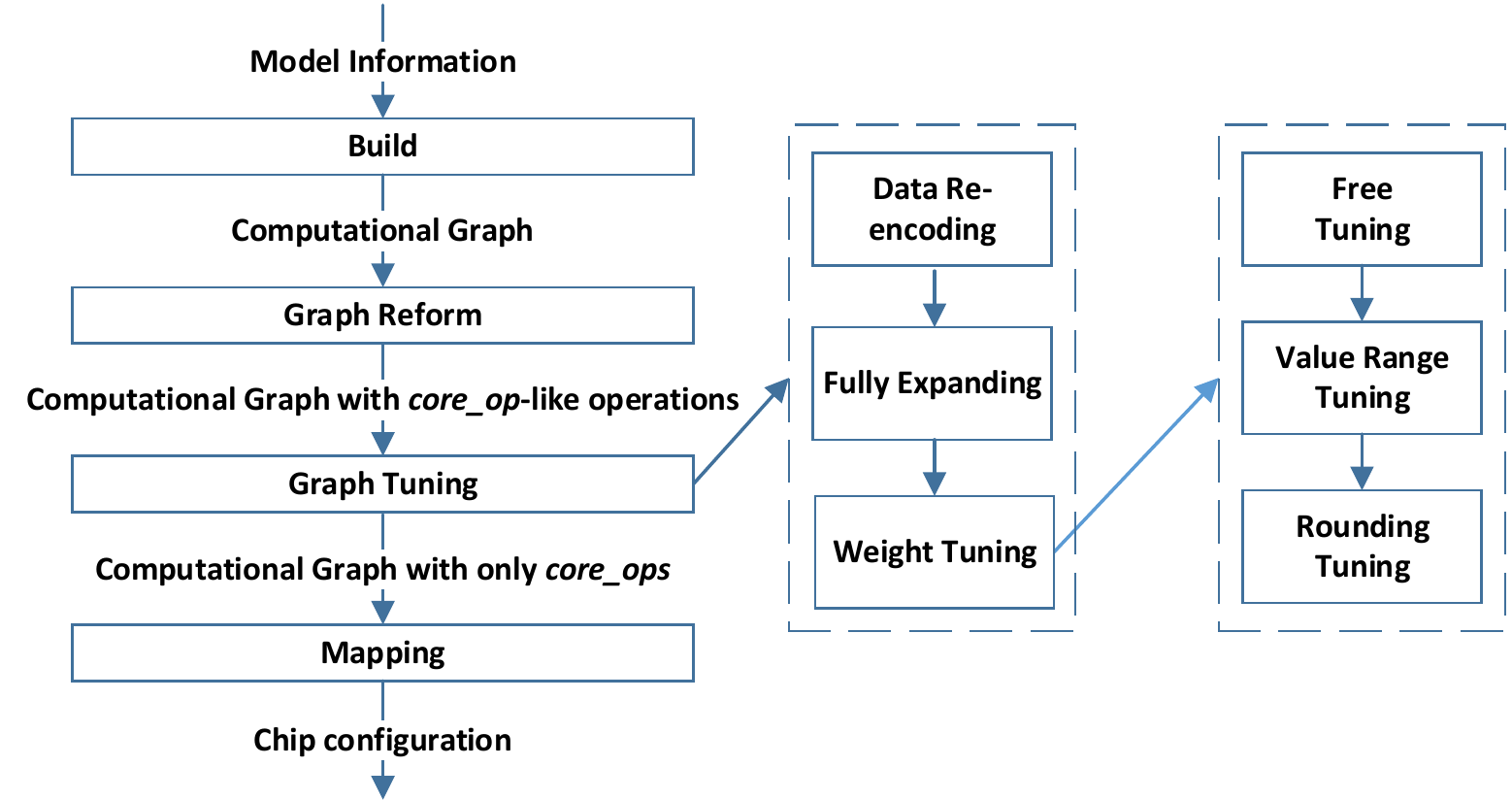}
	\caption{Workflow of our proposal, from the input model to the output for chip. Step `Graph Tuning' contains 3 sub-steps for different hardware restrictions respectively and the third sub-step has 3 fine-tuning phases.}
	\label{fig:workflow}
\end{figure}

The proposed workflow involves 4 steps as shown in Figure~\ref{fig:workflow}.

 \begin{figure}[t]
	\centering
	\includegraphics[width=0.5\textwidth]{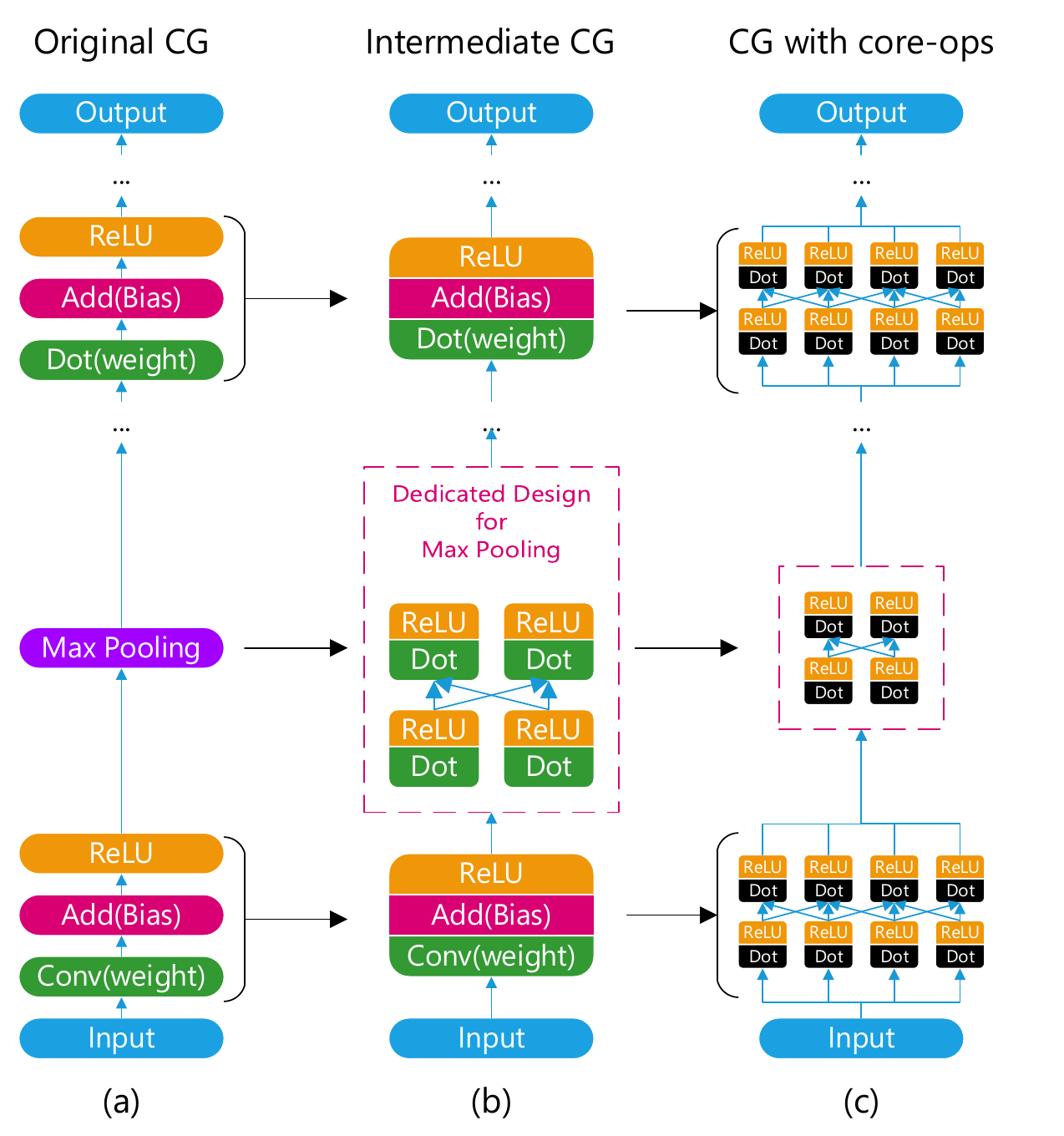}
	\caption{A transformation example}
	\label{fig:example}
\end{figure}

\noindent\textbf{Building CG.}\label{step:parse}
According to the above description, it constructs $G=(V, E)$ based on the input NN's information that includes the trained parameters, network topology, vertex information and training dataset.
An example is shown in Figure~\ref{fig:example}(a).
In addition, we can also get the operation set $F$ supported by the deep learning framework.

\noindent\textbf{Graph Reform.}\label{setp:graph_reform}
It constructs an intermediate CG, $\hat{G}=(\hat{V}, \hat{E})$.
The corresponding operation set $\hat{F}$ contains all operations that are \textit{core-op--like}.
These operations share the similar form as core-ops or can be achieved by core-ops easily, but do not meet those hardware constraints.
Figure~\ref{fig:example}(b) shows an example of the graph with only core-op--like operations.
The detailed transformation from $G$ to $\hat{G}$ is in Section~\ref{sec:graph_reform}.

\noindent\textbf{Graph Tuning.}\label{step:graph_tuning}
In this step, we further transform $\hat{G}$ to $G'$.
Every vertex $\hat{v}\in\hat{V}$ is traversed in a topological ordering of $\hat{G}$ to form corresponding vertices in $G'$.
As shown in Figure~\ref{fig:workflow}, we have multiple sub-steps for the transformation of each vertex $\hat{v}$.
We can also transform a subgraph with multiple vertices as a whole.
We transform the whole graph part by part to have a better convergence since smaller graph are easier to be approximated.
It is where we employ the \textit{divide-and-conquer} principle.
The sub-steps are as following.
\begin{itemize}
    \item \keyword{Data Re-encoding.}
    Re-encode I/O data on each edge of the subgraph to solve the precision problem of I/O data.
    This sub-step is where we employ the \textit{trade-scale-for-capability} principle.
    It enlarges the computation scale, but does not change the operation type, each vertex is still a core-op--like operation.
	\item \keyword{Fully Expanding.}
    Since core-op--like operations are easy to be implemented with core-ops, in this sub-step, we fully expand each core-op--like operation with multiple core-op operations to solve the limitation on the computation scale.
	After this sub-step, the subgraph only contains core-ops.
    \item \keyword{Weight Tuning.}
    This step aims to fine-tune the weight matrices of core-ops in the subgraph to minimize transformation error, under the premise of satisfying the hardware weight precision.
    As shown in Figure~\ref{fig:workflow}, we also introduce three phases of fine-tuning to make it more reliable and easier to converge.
    It is also where we employ the \textit{divide-and-conquer} principle.
 \end{itemize}
Figure~\ref{fig:example}(c) shows an example of the transformed graph $G'$ with only core-ops.
Detailed transformation are in Section~\ref{sec:graph_tuning}.

\noindent\textbf{Mapping.}
Now we have built an equivalent Graph $G'$ with only core-ops that meet hardware constraints, which will be mapped onto the target hardware efficiently.
This part highly depends on the hardware's interconnection implementation.
In Section~\ref{sec:mapping} we will introduce the basic principle of mapping the hardware execution model onto target chips.
 
\subsection{Graph Reform}\label{sec:graph_reform}
In this step, we need to transform a CG $G$ into a graph $\hat{G}$ with only core-op--like operations.
The operation set $\hat{F}$ includes all operations that could be combined by core-ops in $F'$ without any precision constraints.
For example, the corresponding core-op--like operations for dot--core-op are all operations that in the form of weighted sum with activation function, denoted as \textit{dot-like} operations.

To replace all vertices represented with $F$ into operations in $\hat{F}$, we take the following three steps in order.
\begin{textcircleds}
	\item First, we find all subgraphs that match the computation of any $\hat{f}\in\hat{F}$, and replace them with $\hat{f}$.
    For example, in Figure~\ref{fig:example}(b), we merge the dot-product, add-bias and ReLU operations into one operation, which is a dot-like operation.
    \item Then, we can also have some customized mapping from a subgraph in $G$ into a subgraph formed of operations in $\hat{F}$, and apply these dedicated designs here.
    For example, max-pooling operation can be built with max functions.
    A simple max function with two operands can be achieved with as Equation~\ref{eq:max}, which includes multiple dot-like operations.
    \begin{equation}\label{eq:max}
		\begin{split}
    		\max(a, b) = & \frac{1}{2}[\relu(a+b)+\relu(a-b)\\
        			 + & \relu(-a+b)+\relu(-a-b)]\
    	\end{split}
	\end{equation}
    We can use the max function with two operands to form max function with more operands.
    \item Finally, for the left operations in $G$, we provide a default transformation strategy: we use multiple dot-like operations to form MLPs to approximate them since MLP is proved to be an universal approximator~\cite{hornik1989multilayer}.
\end{textcircleds}
After the transformation, we form an graph $\hat{G}$ with only core-op--like operations.

\begin{figure*}[!htbp]
	\centering
	\includegraphics[width=1\textwidth]{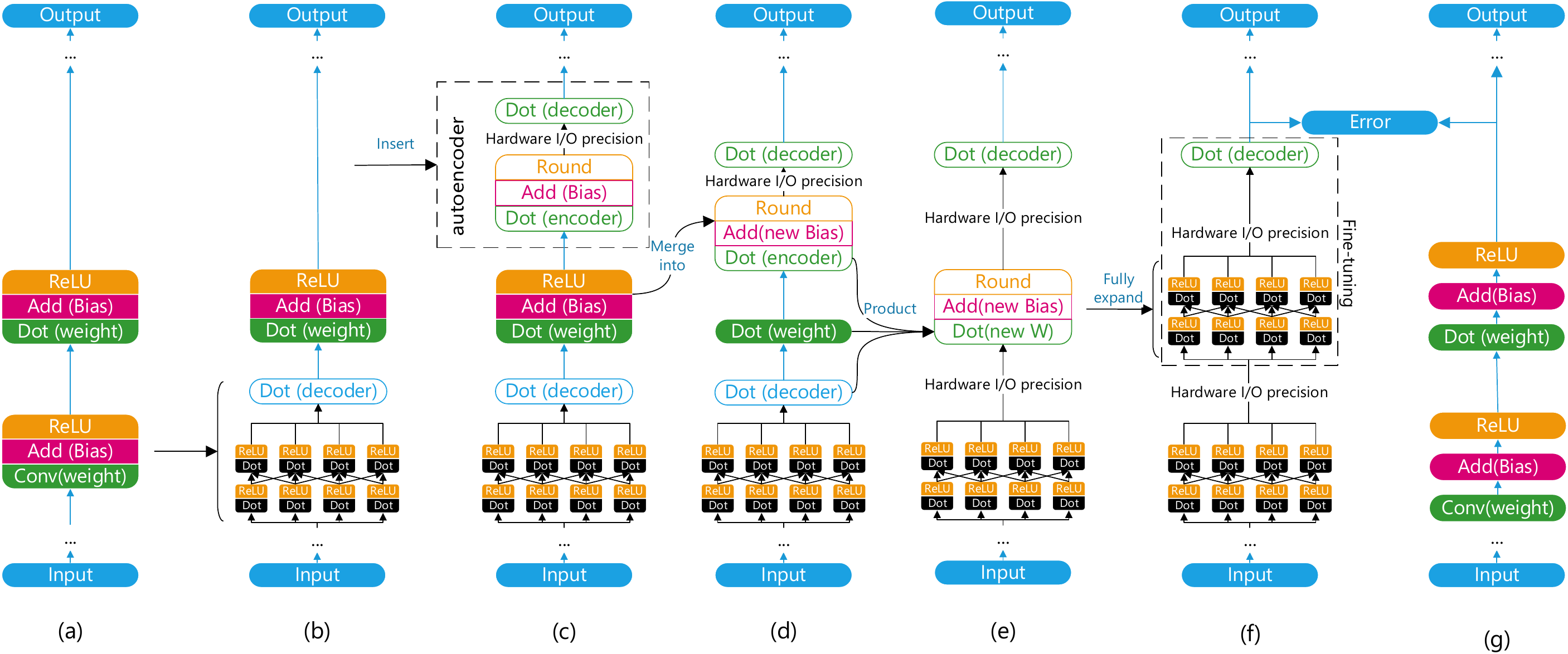}
	\caption{
		Graph tuning of one vertex.
		(a) $G'$ with only \emph{core\_op}-like operations.
		(b) `Graph Tuning' has been performed against all previous vertices.
		(c) Insert an `autoencoder' after the current vertex.
		(d) Merge the bias and activation of the current vertex into the hidden layer of the autoencoder.
		(e) Multiply the three weight matrices together.
		(f) Fully expand the vertices with \emph{core\_ops}.
		(g) The original CG, $G$. The current subgraph (including the decoder of the current vertex) is fine-tuned to approach the corresponding vertex of $G$, with output of the previous tuned vertex as input.
		Repeat (b) to (g) for the rest vertices.
	}
	\label{fig:graph_tuning}
\end{figure*}

\subsection{Graph Tuning}\label{sec:graph_tuning}

In this step, we transform the intermediate graph $\hat{G}=(\hat{V},\hat{E})$ to the hardware execution model $G'=(V',E')$.
We use the original graph $G$ to supervise the fine-tuning progress of the generated $G'$.
The graph $G$ can provide not only labels of the output but also supervised signals of all intermediate data between operations.
Each edge $e\in E$ can provide supervised signal for graph $G'$.
Thus, we can split $\hat{G}$ into parts, and transform it into $G'$ part by part.
To do so, first we find the edges $\hat{e}\in\hat{E}$ that correspond to the edges $e\in E$, and use them to split the graph $\hat{G}$.
Then, we perform the following steps against each part one by one in a topological ordering of $\hat{G}$.
Note that, we can also transform multiple adjacent parts as a whole each time.

\bulletkeyword{Data Re-encoding.}
Since the original model $G$ usually use floating-point numbers for computation and data representation, directly rounding data to hardware I/O precision may distort and lost information.
Thus, this sub-step aims to re-encode the input and output data of each vertex.
To encode a floating-point vector with a low-precision vector, we employ an autoencoder to get the low-precision representation.

An autoencoder is an NN with one hidden layer that can learn a constrained representation of a set of input data.
The output layer has the same dimension as the input layer and is trained to approach the input:
the computation from input to the hidden layer is encoding input data to the hidden layer's representation, while the computation from the hidden layer to output is decoding.

Here as we use it to represent the original floating-point vector with the hardware--I/O-precision vector, it is reasonable that the neuron number of the hidden layer may be greater than that of the input/output layer;
the specific value can be configured manually.
Usually, a large hidden layer will improve the accuracy, but consume more resources.
The tradeoff is evaluated in Section~\ref{sec:evaluation}.

For clarity, we take a vertex of dot-like operation as the example to describe the process.

As shown in Figure~\ref{fig:graph_tuning}(a)(b)(c), we add an autoencoder after the current vertex.
The activation function of its hidden layer is a round operation, which rounds the output to hardware I/O precision.
In addition, the round operation not only quantizes the output to low-precision, but also forces the output to be positive, which provides the non-linearity as the widely-used ReLU function.

The initial weight parameters for the encoder and decoder of the autoencoder are set as following.

For the input vector $X=\{x_1, \ldots, x_n\}$ (i.e. the output of the previous ReLU, as illustrated by Figure~\ref{fig:graph_tuning}(c)), we build an autoencoder network with a hidden layer of $m\times n$ neurons, where $m$ is a configurable integer, denoted as \textit{re-encoding factor}.
Suppose the $n$ dimensions of $X$ are independent and equally important, then each $x$ has $m$ hidden neurons to represent its low precision encoding.
Thus, we initialize the connections from $x$ to other hidden neurons to zero.
A hidden neuron linearly scales the corresponding $x$ to $w^{(e)}x+b^{(e)}$, where $w^{(e)}$ and $b^{(e)}$ are the weight and bias term of the encoder respectively, and then rounds it to the I/O precision, $N$-bit integer ($\{0, \ldots, 2^N-1\}$).
Any $x < \frac{-b^{(e)}}{w^{(e)}}$ will be rounded to $0$ and any $x > \frac{2^N-1-b^{(e)}}{w^{(e)}}$ will be rounded to $2^N-1$.
Thus, one hidden neuron can well represent $x \in [\frac{-b^{(e)}}{w^{(e)}}, \frac{-2^N-1-b^{(e)}}{w^{(e)}}]$.

Now we have $m$ hidden neurons for each dimension;
thus the best way to represent $x \in [0, x_{max}]$ (the output of ReLU is positive) with the $m$ hidden neurons is to divide the data range into $m$ adjacent and non-overlapping intervals, each of which corresponds to one hidden neuron.
Namely, we properly initialize $w^{(e)}_i$ and $b^{(e)}_i$ for the the encoder of the $i$-th neuron of the $m$ to adapt $[\frac{-b^{(e)}_i}{w^{(e)}_i}, \frac{2^N-1-b^{(e)}_i}{w^{(e)}_i}]$ to the corresponding interval $[\frac{ix_{max}}{m}, \frac{(i+1)x_{max}}{m}]$.
Thus $w^{(e)}_i = \frac{(2^N-1)m}{x_{max}}$ and $b^{(e)}_i = -(2^N-1)i$.

Accordingly, to decode and restore $x$, the decoder should scale the data back.
Thus, its weight matrix is set to $w^{(d)}_i = \frac{1}{w^{(e) _i}}=\frac{x_{max}}{(2^N-1)m}$.

Note that, if input $x < 0$ (outside of the encoded interval $[0, x_{max}]$), the initialized autoencoder will always return 0.
It means that the autoencoder can also perform ReLU operation. Therefore, we could remove the redundant ReLU operation of the current vertex.
Moreover, as shown in Figure~\ref{fig:graph_tuning}(d), the bias term of the current vertex could also be encoded by the dot-product operation and merged into the bias term of encoder $b^{(e)}_i$.
Namely, the new bias term of encoder becomes $b^{(e)}_i + w^{(e)}_ib$.

Finally, as shown in Figure~\ref{fig:graph_tuning}(e), the decoder of the previous vertex, the dot-product operation and the encoder of the current vertex can be merged as one dot-product operation, whose weight matrix is the product of the three's matrices.

Till now, the input and output of the current vertex have been constrained to hardware I/O precision.

For vertices of convolution plus activation function, the process is similar.
Instead of using dot-product operation as encoder and decoder, we use convolution instead, and the three convolutions can be merged into one as well.
The initialization is also similar: the hidden layer has $m$ channels for each input channel, and only the center value of the encoder/decoder kernel is set to non-zero.

Owing to this step, we solve the limitation problem of I/O precision.
In this step, we only change the computation scale of the operations in $\hat{G}$.

\bulletkeyword{Fully Expanding.}
In this step, we will turn $\hat{G}$ into $G'$.
Since the core-op--like operations $\hat{f}\in\hat{F}$ can be combined by core-ops in $F'$, we expand all operations in $\hat{G}$ into individual subgraphs consisting of operations in $F'$ to form the graph $G'$.


Take dot-like operations as an example.
Dot-like operations can be represented as a fully-connected layer.
As shown in Figure~\ref{fig:graph_tuning}(f),
to support the fully-connected layer of any scale with dot--core-ops, we use two layers of dot--core-ops to construct an equivalent graph.
The first layer is a computation layer. The original dot-like operations are divided into smaller blocks and are performed with many dot--core-ops.
The second layer is a reduce layer, which gathers result from the former to output.

The division is straight:
\begin{textcircleds*}
	\item Divide the weight matrix into small sub-matrices that satisfy the hardware limitation on scale;
	each is held by a dot--core-op of the first layer.
	\item Divide the input vector into sub-vectors and transfer each sub-vector to the corresponding sub-matrices(\emph{core\_ops}) at the same horizontal position and
	\item gather results from the same column by the reduce layer.
\end{textcircleds*}

Regarding all dot-like operations as fully-connected layers are sometimes very inefficient. We can have dedicated division according to its connection pattern.

For example, for a convolutional case (suppose a kernel of size $k \times k$ convolves a $W \times H$ image from $m$ channels to $n$ channels), the $n$ channels of one pixel in the output side are fully connected to the $m$ channels of $k \times k$ corresponding pixels in the input side.
This forms a small-scale vector-matrix multiplication of size $(m \times k^2) \times n$.
There are $W\times H$ such small operations in the convolution case.
Each input should be transferred to $k^2$ such small operations, while reduction is needless.
If such a small operation is still too large for a dot--core-op, we can divide the operation as the fully-connected--layer case does.

If there are some dot--core-ops that are not fully used, we can distribute them onto one physical PE to reduce resource consumption during the mapping step.

Till now, the computation of the current subgraph of $\hat{G}$ has been transformed to a subgraph of $G'$, which consists of core-ops $f'\in F'$.
Next, the weight matrices of the core-ops will be fine-tuned.

\bulletkeyword{Weight Tuning.}
In this step, we will fine-tune the parameters to make the generated subgraph of $G'$ approximately equal to the corresponding subgraph of $G$.

As shown in Figure~\ref{fig:graph_tuning}(g)(f), we use the corresponding supervised signal from graph $G$ to fine-tune current subgraph of $G'$.
The input to the current subgraph is from the output of previous transformed subgraphs instead of the corresponding supervised signal from the graph $G$.
Thus, the transformation of current subgraph will consider the error from previous transformed subgraphs, which can avoid error accumulation.
The output of previous transformed subgraph can be generated on demand or cached in advance to improve the transformation speed.

 

We will consider the hardware constraints on weight parameters in this step.
Specifically, target hardware usually puts strict constraints on weight storage since it occupies most of the hardware resources.
Here we present a formal description of the constraints on the weight matrix $W$:
the value of each element $W_{ij}$ should be assigned dependently from a finite set $S$.
$S$ is either a fixed set or a configurable set $S^P$ with parameter(s) $P$.
Three kinds of typical weight encoding methods, which have been widely used by real NN chips, are presented as following (in all cases, the size of $S$ is $2^N$):
\begin{itemize}
	\item \textbf{Dynamic fixed-point}: $S^P=\{\frac{-2^{N-1}}{2^P}, \ldots, \frac{0}{2^P}, \ldots, \frac{2^{N-1}-1}{2^P}\}$ where $P$ represents the point position.
	This method is used by DNPU~\cite{DNPU2017}, Strip~\cite{judd2016stripes}, TianJi-ANN~\cite{11}, etc.
	\item \textbf{Fraction encoding}: $S^P=\{\frac{-2^{N-1}}{P}, \ldots, \frac{0}{P}, \ldots, \frac{2^{N-1}-1}{P}\}$, where $P$ is the threshold of the spiking neuron or the scale factor of the result. It is used by PRIME~\cite{7}, and TianJi-SNN~\cite{11}.
	\item \textbf{Weight sharing}: $S^{P_1, \ldots, P_{2^N-1}}=\{0, P_1, \ldots, P_{2^N-1}\}$, used by EIE~\cite{han2016eie}.
\end{itemize}

Without loss of generality, suppose the floating-point parameter $W_{ij}$ is rounded to the $k_{ij}$-th element in $S^P$, denoted as $S^P_{k_{ij}}$.
This step aims to find the best $P$ and to set $k_{ij}$ for each element in the weight matrix properly to minimize the transformation error.
It is similar to \textit{weight quantization} of network compression.
Our contribution is that we generalize it to typical hardware cases and introduce several fine-tuning phases to deal with different parameter-setting issues separately.

For a subgraph, three fine-tuning phases are taken in order:
The first is to reduce the initialization error.
The second is to determine the best value range of weight matrix (i.e. to choose the best $P$) and the last is to determine the best value from $S^P$ for each element (i.e. to choose the best $k_{ij}$).
Each phase gets parameters from the previous one and fine-tunes them under certain constraints.

\bulletkeyword{Free Tuning.}
In previous steps, we use the parameters in the original graph $G$ to initialize those parameters in the generated graph $G'$.
However, some methods, including autoencoder and the MLP-based unsupported-function handling, introduce transformation errors.
In addition, activation functions used by $G$ may be different from the hardware counterpart, which also makes the initialization inaccurate.
In addition, previous transformed subgraphs also have errors.
Therefore, some fine-tuning phases have to be taken to minimize the error, under the premise of satisfying the hardware constraints on weight precision.

Thus we first fine-tune the subgraph of $G'$ without any constraint on weight precision to reduce any existing error.
In this procedure, all parameters and signals are processed as floating-point numbers, while the hardware activation function is used.

\bulletkeyword{Value-Range Tuning.}
Now the precision constraint on weight is introduced.
Accordingly, we need to choose the best value-range of the weight matrix (namely, the best $P$).
Apparently, we will minimize $J(k, P)=\sum_{ij}(W_{ij}-S^P_{k_{ij}})^2$, which can be achieved by an iterative expectation\-maximization (EM) algorithm:
\begin{itemize}
	\item E-step: fix the current parameter $P^{(t)}$ and calculate $k_{ij}^{(t)} = \arg\min{J(k|P^{(t)})}$.
	\item M-step: fix $k_{ij}^{(t)}$ and calculate $P^{(t+1)}=\arg\min{J(P|k^{(t)})}$.
\end{itemize}

Then we replace $W_{ij}$ with $S^P_{k_{ij}}$ where $k_{ij}$ is fixed and $P$ is the parameter.

After the initialization, we fine-tune the subgraph to optimize $P$.
During this process, we maintain the precision of $W_{ij}$ first and then round it to $P_{k_{ij}}$ at every time $P$ is updated.

Further, for the weight sharing case mentioned above, the EM algorithm is just reduced to the k-means algorithm.
If $S^P$ is a fixed set without any configurable parameter, we can omit this phase.

\bulletkeyword{Rounding Tuning.}
The data-range set of weight value $S^P$ is fixed now.
This procedure adjusts each weight matrix element to a proper element in this set.
In another word, it aims to choose the best index $k_{ij}$ for $W_{ij}$.
During the fine-tuning progress, parameters are stored as floating point number.
In the forward phase, any parameter is rounded to the closest element in $S^P$.
While during the backward phase, floating-point number is used to update $W_{ij}$.
This mechanism is also employed by the above \textit{Value-Range Tuning} phase if $P$ can be set from a discrete set.


After processing all the subgraphs, we have transformed the original model $G$ into an equivalent hardware execution model $G'$ that satisfies all the constraint conditions.

\subsection{Mapping}\label{sec:mapping}
The generated graph $G'$ will be deployed on the target hardware efficiently, which is a hardware-specific problem.
Thus, we give the optimization principle here.

For NN chips that bind the neural computation and storage in the \textit{physical cores} (it is called the \textit{weight stationary} computing mode, classified by~\cite{16}), this is a mapping problem to assign core-ops to physical cores. Moreover, several core-ops that are not fully used can also be distributed onto one physical core, as long as there are no data conflicts.

For chips whose physical cores are computing engines with flexible memory access paths to weight storage (usually work in the time division multiplex mode), it is a mapping and scheduling problem to schedule each core-op's task onto physical cores.
Multiple core-ops could be mapped onto one core to increase resource utilization.

As we can get data dependencies and communication patterns between core-ops through the transformed graph, we could use these information to optimize the mapping or scheduling to minimize transmission overhead, e.g. putting densely-communicating cores close.
TrueNorth has designed such an optimized mapping strategy~\cite{38}.

Moreover, for those core-ops sharing weights (e.g. convolution vertices can be fully expanded to a lot of core-ops sharing the same weight matrix), we could map (or schedule) them to the same physical core to reduce data movement.

\subsection{Others}\label{sec:specification}
\subsubsection{SNN Models}\label{sec:snn}
SNN, called the third generation of ANN, is a widely-used abstraction of biological neural networks.
In addition to neuronal and synaptic states that traditional ANN has featured, it incorporates the timing of the arrival of inputs (called spikes) into the operating model to yield higher biological reality.

SNNs of rate coding can emulate ANNs.
The spike count in a given time window can represent a numerical value within a certain range, like a traditional ANN does.
Accordingly, the input of a synapse is a spike sequence of certain firing rate from the pre-neuron.
After synapse computation, it is converted into the sum of currents that will be computed by the post-neuron.
For those widely-used SNN models, the functions of their synapse and neuron computations usually own good continuity and are derivable in rate coding domain.
Therefore, the popular SGD method can be used for training SNN:
several recent studies~\cite{46, 49} have used the stochastic gradient decent (SGD) algorithm to train SNNs directly or indirectly and achieved the state-of-the-art results for some object recognition tasks.

As our workflow is not dependent on the concrete NN type (ANN or rate-coding SNN), it can support SNN hardware and SNN models, too.
For SNN models, the training data is the firing rate of each neuron.

\subsubsection{RNN Models}
RNN is an NN with some cycle(s).
We could transform and fine-tune each operation inside an RNN as normal, and add an additional step to fine-tune the entire RNN after that.

\begin{table*}[!t]
	\centering
	\begin{tabular}{*{8}{l}}
	\hline
	\textbf{NN model} & \textbf{Chip} & \textbf{Weight Encoding} & \textbf{Weight} & \textbf{I/O} & \textbf{Re-encoding} & \textbf{Top1 Accuracy}\\
    & & & \textbf{Precision} & \textbf{Precision} & \textbf{Factor} & \textbf{(Accuracy Drop)}\\
    \hline\hline
  	\textbf{MNIST-MLP} & & Floating-point & & & & 98.2\%\\\hline
	MNIST-MLP & TianJi-ANN & Dynamic fixed-point & 8-bit & 8-bit & $1\times$ & 98.15\%(-0.05\%)\\
	MNIST-MLP & TianJi-SNN & Fraction encoding & 8-bit & 1-bit & $2\times$ & 96.59\%(-1.61\%)\\
	MNIST-MLP & TianJi-SNN & Fraction encoding & 8-bit & 2-bit & $2\times$ & 97.63\%(-0.57\%)\\
	MNIST-MLP & PRIME & Fraction encoding & 8-bit & 6-bit & $1\times$ & 98.14\%(-0.06\%)\\
    \hline
    \textbf{LetNet-5} & & Floating-point & & & & 99.1\%\\\hline
	LeNet-5 & TianJi-ANN & Dynamic fixed-point & 8-bit & 8-bit & $1\times$ & 99.08\%(-0.02\%)\\
	LeNet-5 & PRIME & Fraction encoding & 8-bit & 6-bit & $1\times$ & 99.01\%(-0.09\%)\\
    \hline
    \textbf{CIFAR10-VGG17} & & Floating-point & & & & 84.64\%\\\hline
	CIFAR10-VGG17 & TianJi-ANN & Dynamic fixed-point & 8-bit & 8-bit & $1\times$ & 84.02\%(-0.62\%)\\
    CIFAR10-VGG17 & PRIME & Fraction encoding & 8-bit & 6-bit & $1\times$ & 83.57\%(-1.07\%)\\
    \hline
    \textbf{ImageNet-AlexNet} & & Floating-point & & & & 57.4\%\\\hline
    ImageNet-AlexNet & TianJi-ANN & Dynamic fixed-point & 8-bit & 8-bit & $1\times$ & 56.9\%(-0.5\%)\\
	ImageNet-AlexNet & PRIME & Fraction encoding & 8-bit & 6-bit & $1\times$ & 55.2\%(-2.2\%)\\
    ImageNet-AlexNet & PRIME & Fraction encoding & 8-bit & 6-bit & $4\times$ & 57.0\%(-0.4\%)\\
    \hline
    \textbf{ImageNet-VGG16} & & Floating-pint & & & & 70.5\%\\\hline
	ImageNet-VGG16 & TianJi-ANN & Dynamic fixed-point & 8-bit & 8-bit & $1\times$ & 69.6\%(-0.9\%)\\
	ImageNet-VGG16 & PRIME & Fraction encoding & 8-bit & 6-bit & $1\times$ & 68.2\%(-2.3\%)\\
    ImageNet-VGG16 & PRIME & Fraction encoding & 8-bit & 6-bit & $4\times$ & 69.5\%(-1.0\%)\\
    \hline
	\end{tabular}
	\caption{Accuracy for NNs under different restrictions}
	\label{tab:nn_model_transform}
\end{table*}

\section{Implementation and Evaluation}\label{sec:evaluation}
\subsection{Implementation}
We have implemented the tool to support different hardware constraints, including those of TianJi~\cite{11} and PRIME~\cite{7}.

TianJi is fabricated with 120nm CMOS technology. The running frequency is 100MHz and the total dynamic power consumption is 120mW. TianJi chip supports both ANN and SNN modes.
The numerical accuracy of weight value is 8-bit fixed-point and the scale of vector-matrix-multiplication is $256 \times 256$.
For ANN mode, the I/O precision is 8-bit that is cut from the 24-bit internal computation output; the cut range is configurable, thus its weight encoding strategy is dynamic fixed-point.
For SNN mode, the minimal I/O precision is 1-bit, which can be extended to $n$-bit with $2^n$ cycles as the sampling window (as described in Section~\ref{sec:snn}). The neuron model is a simplified LIF neuron model with a constant leakage and a threshold for firing; the weight encoding method can be viewed as fraction encoding.
PRIME~\cite{7} is a memristor-based PIM architecture for ANN.
The weight precision is 8-bit and the I/O precision is 6-bit. The scale of vector-matrix-multiplication is $256 \times 256$, too. The output range can be configured with an amplifier; thus its weight encoding can also be viewed as fraction encoding.

Quite a few NN applications, including an MLP for MNIST dataset (784-100-10 structure, $98.2\%$ accuracy of full precision), LeNet-5~\cite{LeNet} for MNIST dataset ($99.1\%$ accuracy), a CNN~\cite{mishkin2015all} for CIFAR-10 dataset ($84.64\%$ accuracy \footnote{As described by ~\cite{mishkin2015all}, with some special initialization method, the CNN accuracy can exceed 90\%. Here we ignore it for simplicity, which does not affect our evaluation.}), AlexNet~\cite{AlexNet} and VGG16~\cite{VGG16} for ImageNet, have been respectively transformed and then deployed onto TianJi~\cite{11} and PRIME~\cite{7} to show the validation.
The first three networks are trained by Theano~\cite{2016arXiv160502688full}.
Parameters of the next two CNNs for ImageNet are extracted from trained models of the Caffe Model Zoo directly.
The inference accuracies of full precision are given in Table~\ref{tab:nn_model_transform}.

Without loss of generality, we take the mapping of the LeNet-5 for MNIST onto the TianJi system as an example.

One TianJi chip contains 6 cores connected by a $2 \times 3$ mesh NoC;
each core supports 256 simplified LIF neurons and works in the weight stationary mode.
The main body of a TianJi system is a Printed Circuit Board (PCB) including 16 chips.
On the whole, all of the cores form a $12 \times 8$ 2D-mesh network, and there is a great gap between the delay/bandwidth of intra-/inter-chip communications.

The transformed CG consists of 497 TianJi's dot--core-op.
Taking into account the weight reuse of convolution, 19 physical cores are actually occupied.
We use the heuristic Kernighan-Lin (KL) partitioning algorithm for the mapping problem.
It abstracts the mapping as a graph partition problem and tries to minimize communications across partition boundaries.
Take the bipartition as an example: the input to the algorithm is a graph;
the weight of each edge is the communication delay.
The goal is to partition the graph into two disjoint subset $A$ and $B$ of equal size, in a way that minimizes the communication cost of the subset of edges that cross from $A$ to $B$.

First, a randomly generated initial mapping distribution is given.
Second, the KL algorithm bi-partitions the mapped cores repeatedly till only the closest two cores are left in any of the final partition in the 2D-mesh. During this phase, partitions that minimize the communication cost between cores are remained; here the cost of an edge across boundary refers to its weight multiplied by the number of transmissions, as we can get the communication statistics from the transformed CG, including those information about the reused cores.

\subsection{Evaluation}

The inference accuracies after transformation for TianJi and PRIME are given in Table~\ref{tab:nn_model_transform}.
This table also shows the re-encoding factor, which indicate the number of hidden units for autoencoders, for each case.

We can see that the transformation errors introduced are very limited, for all NN models and chips tested, we can achieve less than 2\% accuracy drop.
For most cases, we only use $1\times$ hidden unit to re-encode data, which means that the NN scale is not changed.
For the two TianJi-SNN cases, as the I/O constraints are very strict, we use $2\times$ hidden units, thus $4\times$ PEs will be used.
For ImageNet cases on PRIME, if we only use $1\times$ hidden units, the accuracy drop will be over 2\%.
With $4\times$ hidden units, we can reduce the accuracy drop to less than 1\%.
We can always achieve better accuracy with more hardware resources employed.


This toolchain also improves the development efficiency remarkably. For example, it can transform AlexNet in about 1 hour, while training it from scratch will take about 3\textasciitilde4 days. Specially, training the whole NN requires millions of iterations to converge, while our method only costs thousands of iterations to converge for each step, that is, takes 5 to 10 minutes to transform one layer. The reason lies in that, after partitioning, we fine-tune each unit one by one; each one is well initialized and much easier to converge than training the entire model. All evaluations are completed on a common PC server equipped with one Tesla P100 GPU.

In addition, as the large-scale NNs (e.g. those for ImageNet) cannot be occupied by the TianJi system because of the physical limit of chip capacity (a TianJi system contains 16 chips and a chip can occupy 1536 neurons), those related results are drawn from the cycle-accurate TianJi chip simulator.
\subsubsection{Accuracy vs. Fine-Tuning Granularity}
We conduct experiments to explore the relationship between accuracy and fine-tuning granularity.
In another word, in the step of \textit{Graph tuning}, we can fine-tune one or more successive parts in $G'$ simultaneously, even fine-tune the entire model as a whole.
It looks like that increasing the fine-tuning scale per time will increase the search space, which may lead to a lower error rate but consume more computations and more time to converge.
However, results show that coarse grained fine-tuning does not result in improved accuracy.
For example, for CIFAR10-CNN, the inference accuracy (the I/O precision is 7 bits and the weight precision is 4 bits) is 83.07\% as the fine-tuning-granularity is two subgraphs (the accuracy is 83.14\% as the granularity is one).
If we fine-tune the whole NN together, the accuracy is just 83.24\%.
Thus, one by one fine-tuning is an optimal strategy, which also means that the problem of error accumulation has been well solved.
Unless specifically noted, the fine-tuning granularity is just one.

\subsubsection{Accuracy vs. Resource Consumption}
From the transformation workflow, we can see that the step of \textit{Data Re-encoding} may introduce the most additional resource overhead:
When the neuron number of the hidden layer of autoencoder is $n\times$ as much as that of the input/output layer, the number of crossbar consumed of the whole NN will be $n^2 \times$.
The latter can also be considered as a direct indicator of area consumption and runtime overhead. Thus, we conduct experiments to explore the effect of \textit{autoencoder}.

We use the MLP for MNIST dataset.
Although this network is relative small, its conclusion is general for large-scale NNs because we fine-tune NNs part by part and each part is a small graph.
Table~\ref{tab:nn_model_transform} also shows the improvement for ImageNet on PRIME with a larger re-encoding factor.
We compare the inference accuracies without or with different scales of autoencoder.
For the former, we simply scale and round the I/O signal values to make them suitable for I/O precision.

\begin{figure}[t]
	\centering
	\includegraphics[width=0.48\textwidth]{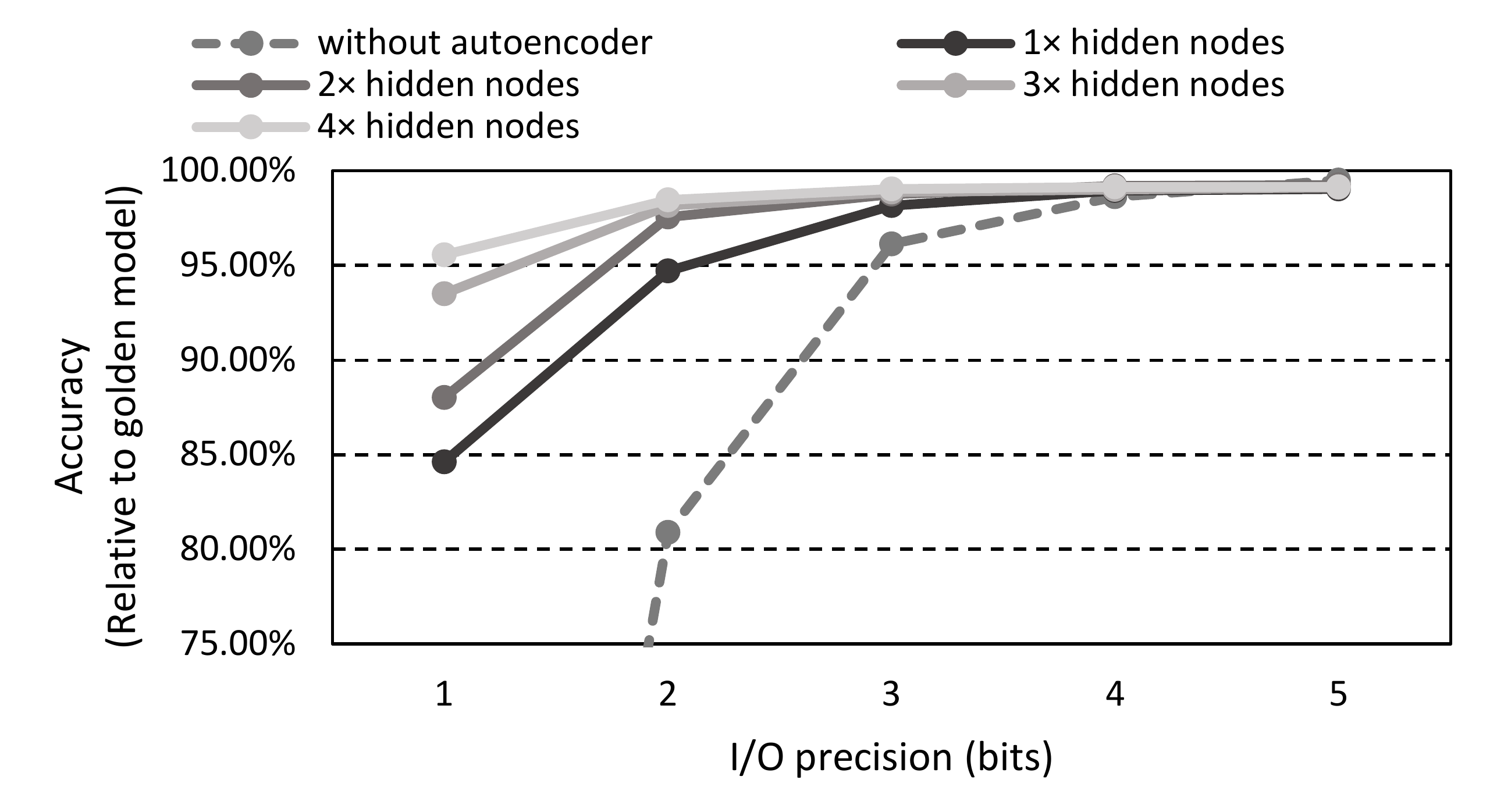}
	\caption{
		Accuracy v.s. I/O precision under different transformation strategies.
	}
	\label{fig:autoencoder_io}
\end{figure}
\begin{figure}[t]
	\centering
	\includegraphics[width=0.48\textwidth]{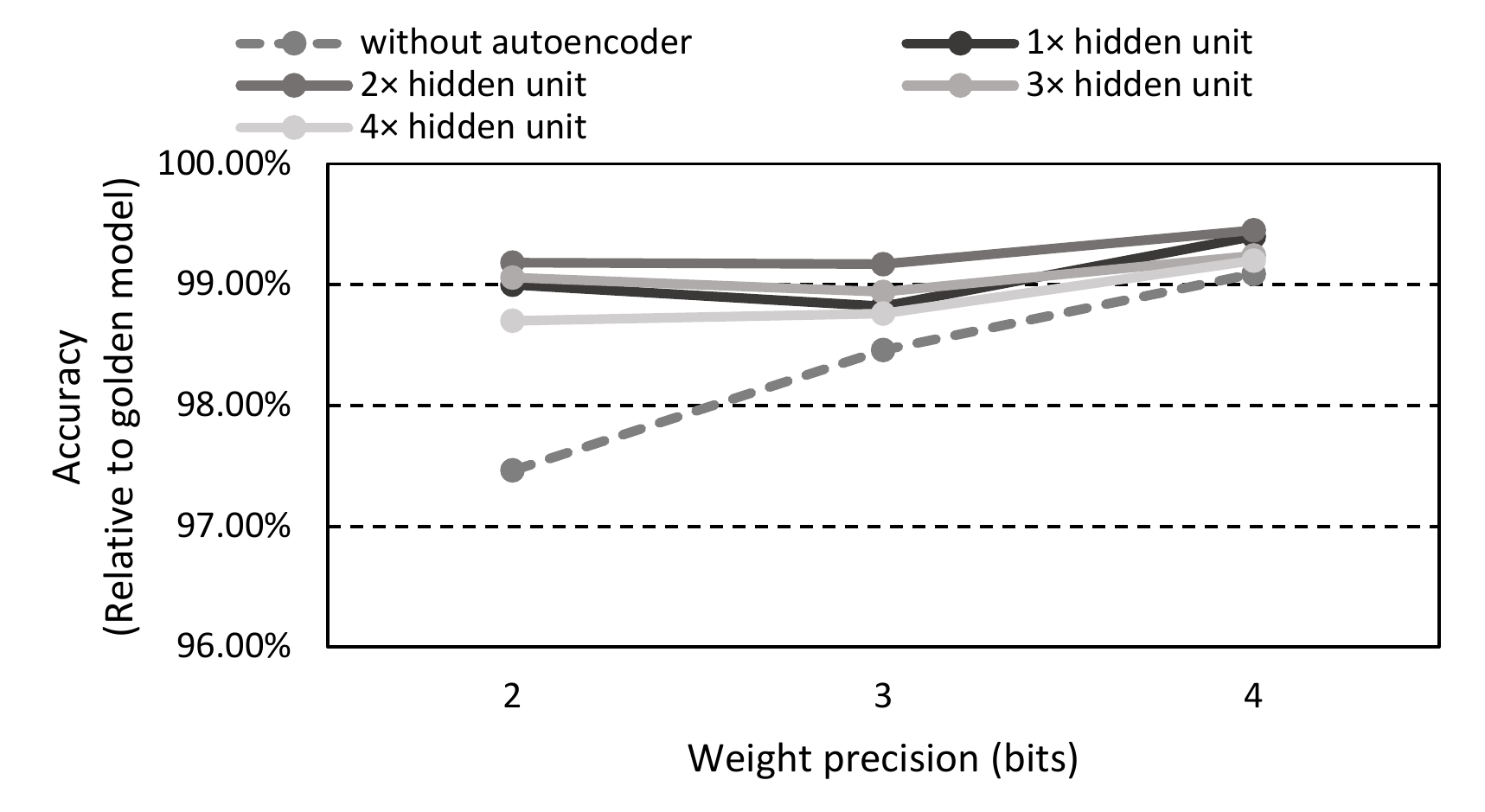}
	\caption{
		Accuracy v.s. weight precision under different transformation strategies.
	}
	\label{fig:autoencoder_weight}
\end{figure}

Figure~\ref{fig:autoencoder_io} (Figure~\ref{fig:autoencoder_weight}) lists the accuracy with different I/O (weight) precisions respectively (after transformation), without or with different scales of autoencoder, while no other constraints are introduced. Results show that this data re-encoding strategy is effective to improve the transformation accuracy under strict constraints.

In Figure~\ref{fig:autoencoder_io}, the accuracy of the transformed network without autoencoder drops significantly when I/O precision is less than 3 bits (different NNs may have different turning points). In contrast, when autoencoder is used (we assume the I/O limitation is only 1-bit and only 1x hidden neurons are used, namely, the most critical case), the accuracy is $84.63\%$ (if no autoencoder, the value is only $13.45\%$). With more hidden neurons, the accuracy continues to rise, which means our method could trade NN scale for capability.

Apparently, increasing the number of hidden neurons can only linearly increase the encoding ability, which is worse than increasing the I/O precision directly because the latter's encoding ability is $\propto(2^n)$. For example, using $2\times$ hidden neurons and 1-bit I/O does consume the same number of I/O ports with that of using $1\times$ hidden nodes and 2-bit I/O; the accuracy of the former is only $88.2\%$ while the latter is $94.71\%$. Thus, it looks like that the hardware had better provide enough I/O precision since rescuing the accuracy by software (using autoencoder) may cost more hardware resources, especially when the hardware I/O precision is less than the turning point. 
Anyway, this is a tradeoff between hardware consumption, software adaption and inference error.

Moreover, as illustrated by Figure~\ref{fig:autoencoder_weight}, autoencoder is also able to rescue the accuracy loss caused by low weight precision. Compared with Figure~\ref{fig:autoencoder_io}, we can see that NNs are more tolerant of low weight precision than low I/O precision, since the latter can cause signal distortion directly. Figure~\ref{fig:autoencoder_weight} also shows that when the weight precision is 2-bit or more, using different scales of autoencoder does not change the accuracy apparently because it has already reached $99\%$.

\subsubsection{Impact of Weight Encoding Methods}
\begin{figure}
	\centering
	\includegraphics[width=0.48\textwidth]{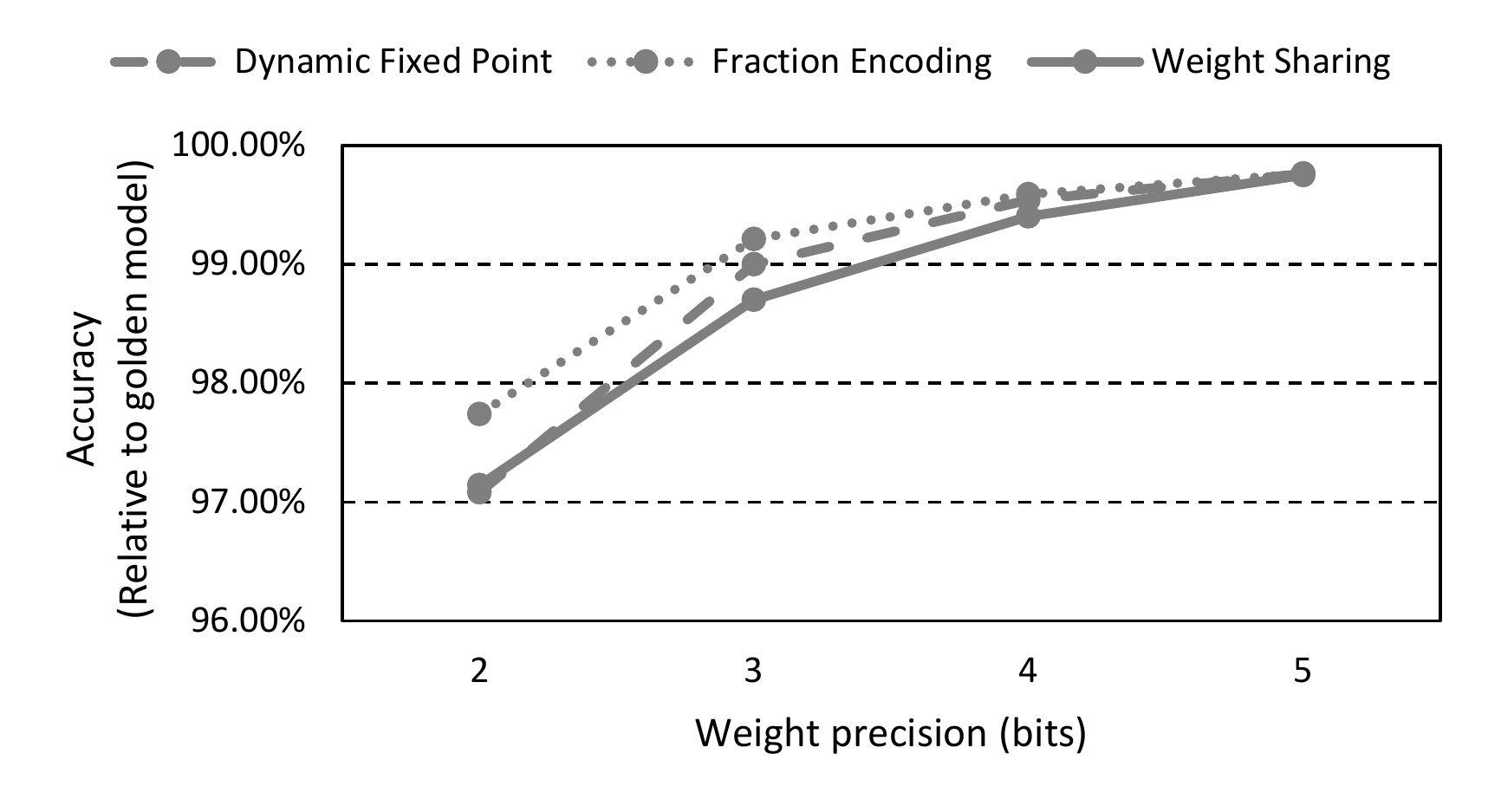}
	\caption{
		Accuracy v.s. weight precision under different weight encoding strategies.
		For the weight sharing case, weight precision means the bit-width of weight indices.
	}
	\label{fig:em_weight}
\end{figure}

We have evaluated the weight tuning algorithm, as well as the three kinds of weight encoding strategies. Figure~\ref{fig:em_weight} shows that weight tuning can set the weight parameters well for the three cases: with the increase of weight precision (any other constraint is not introduced), all of them can reach the upper bound accuracy. In Table~\ref{tab:weight_encoding}, we further give the effect of each phase of the weight tuning step (the weight precision is 2-bit, without any other constraint).

\begin{table}[!t]
	\centering
	\begin{tabular}{|c|c|c|c|}
		\hline
		& \bfseries dynamic & \bfseries fraction  & \bfseries weight sharing\\
        & \bfseries fixed point & \bfseries encoding & \bfseries \\ \hline
		\bfseries I & 81.56\% & 85.60\% & 86.19\%\\ \hline
		\bfseries I+P & 81.56\% & 87.05\% & 89.86\%\\ \hline
		\bfseries I+L & 97.08\% & 97.05\% & 96.31\%\\ \hline
		\bfseries I+P+L & 97.08\% & 97.74\% & 97.14\%\\ \hline
	\end{tabular}
		\caption{
		Accuracy v.s. different weight encoding strategies under 2-bit weight precision.
		I: initialization with EM; P: parameter range fine-tuning (to decide $P$); L: low precision fine-tuning (to decide $k_{ij}$)
	}
	\label{tab:weight_encoding}
\end{table}

With only the EM-based initialization of the value-range tuning phase (I in Table~\ref{tab:weight_encoding}), the accuracy of different weight encoding strategies depends on the latter's flexibility. The accuracy of \textit{weight-sharing} is the highest as it is the most flexible: the precision just limits the bit-width of indices, not weight values. \textit{dynamic-fixed-point} only allows weight values to be scaled by power of 2; thus it is the least flexible. \textit{fraction-encoding} is positioned in the middle.

With the whole value-range tuning phase (including the initialization and fine-tuning, I+P in Table~\ref{tab:weight_encoding}), the accuracy increases a little for both \textit{fraction-encoding} and \textit{weight-sharing} and remains unchanged for \textit{dynamic-fixed-point}. The reason is in the initialization we have already found a good $P$, further fine-tuning $P$ cannot increase the accuracy much, especially for \textit{dynamic-fixed-point} with limited flexibility.

With the EM initialization and the rounding tuning phase (I+L in table~\ref{tab:weight_encoding}), the accuracy increases significantly, which means that NN can easily find suitable parameters from any well-initialized set of weight values.

With all phases employed (I+P+L in table~\ref{tab:weight_encoding}), the accuracy could still increase a little compared with the I+L case, except for \textit{dynamic-fixed-point}.

Anyway, there is no obvious capability difference between the three strategies.

\subsection{Discussion and future work}\label{sec:discussion}
Now, some transformation steps introduce extra layers more than one time, which may exacerbate NN redundancy.
Therefore it is necessary to strike a balance between the possible information loss and hardware-resource consumption.
Anyway, we provide a framework, while the concrete workflow could be customized.
Moreover, the interaction between network compression and transformation is interesting, and we will study it as the future work.

In addition, it is helpful to present insights into future neuromorphic architecture designs:
\begin{squishlist}
	\item It could give design tradeoff between the common computational components and special functional components.
	For some neural functions or layers that are relatively easy (in terms of hardware consumption) to be achieved by common core-ops, it is unreasonable to integrate such a dedicate component on chip.
	If not, a special component is worth to realize.

	\item After transformation, the data flow between core-op is basically determined;
	we can analyze the communication pattern in detail, which is conductive to the balanced distribution of computing and communication resources on chip.

	\item Our solution regards NN inference as the forward process of graph with fixed topology. To some extent, it is suitable for field programmable devices (especially in the aspect of on-chip connection); thus how to combine the device configurability with the flexibility of transformation is an interesting topic.
\end{squishlist}

\section{Related Work}\label{sec:related_work}
There are two types of neural network chips.
The first type focuses on the traditional ANNs: they are custom architectures to accelerate mature artificial intelligence (AI) algorithms, especially machine learning algorithms of DNN.
The second usually supports SNNs to yield higher biological reality.
In order to distinguish, we call the former NN accelerators and the latter neuromorphic chips in the following content.
\subsection{NN compression}\label{sec:nn_compression}
NNs are both computational intensive and memory intensive, making them difficult to deploy on hardware systems with limited resources.
At the same time, there is significant redundancy for deep learning models~\cite{denil2013predicting}.
Therefore, quite a few studies have been carried out to remove the redundancy, which can be divided into three types: \emph{pruning neurons}, \emph{pruning synapses} and \emph{pruning weights}.
Both \emph{weight-quantization}~\cite{vanhoucke2011improving, hwang2014fixed, anwar2015fixed} and \emph{weight sharing}~\cite{chen2015compressing} are pruning weights.
For pruning synapses, deep compression~\cite{han2015deep} prunes the network by retaining only important connections.
Diversity Network~\cite{45} prunes neurons, which selects a subset of diverse neurons and fuses the redundant neurons into the selected ones.

Moreover, there are some research efforts that study extremely compact data representations (including the I/O precision or weight precision or both) for NN computation. Binarized neural networks [1,3] that investigates the use of 1-bit data types for weight and ternary neural networks [4,5] using 2 bits belong to this category, which have achieved comparable accuracies to state-of-the-art full precision networks for some data sets. However, these methods are effective for specific networks, not common development scenarios.
\subsection{NN accelerators}\label{sec:nn_accelerators}
EIE~\cite{han2016eie} extensively employs the above compression techniques and then proposes a dedicated engine to perform inference on the compressed network model.
ESE~\cite{han2016ese} is its follow-up work that implements a speech recognition engine with compressed Long-Short-Term-Memory model on FPGA.

The DianNao chip family is a series of state-of-the-art NN accelerators.
DianNao~\cite{13} designs an accelerator for DNNs and CNNs that exploits data reuse with tiling.
The inherent sharing of weights in CNNs is explored in ShiDianNao~\cite{21}.
PuDianNao~\cite{14} supports seven machine learning algorithms.
DaDianNao~\cite{15} is a custom multi-chip machine-learning architecture.
From the aspect of internal data representation and computation, they support fixed-point computation rather than 32-bit floating-point to reduce hardware cost.
Accordingly, some corresponding retraining or tuning process is needed to adapt software for hardware.
They use a load-store ISA~\cite{liu2016cambricon} to decouple synaptic weight storage from neuron processing logic to avoid the limitation on connection number and improve the utilization of processing logic.
This family also supports compressed, sparse NNs through a dedicated accelerator, Cambricon-X~\cite{zhang2016cambricon}.

Minerva~\cite{27} uses fine-grained data type quantization and dynamic operation pruning to further reduce the resource consumption of DNN hardware.
Strip~\cite{judd2016stripes} relies on bit-serial compute units to enable support for per-layer, dynamically configurable computation precision for DNN.
DNPU~\cite{DNPU2017} supports dynamic fixed-point with online adaption and weight quantization, too.
Other studies include Origami~\cite{22}, Convolution Engine~\cite{26}, RedEye~\cite{28}, NeuroCube~\cite{29}, neuFlow~\cite{31} and quite a few FPGA-based designs~\cite{23, 24, 25, sharma2016high}.

All the above studies are based on the traditional Complementary Metal-Oxide-Semiconductor (CMOS) technology.
Another category is using novel nonvolatile memory devices (usually memristors) to perform neural computations in memory.
PipeLayer~\cite{songpipelayer} is such an accelerator for CNNs that supports both training and inference, while PRIME~\cite{7} and ISAAC~\cite{18} are for inference.
Other work includes stand-along accelerators~\cite{1, hu2016dot, chen2015optimized}, co-processor~\cite{li2013memristor-based} and many-core or NoC~\cite{liu2015reno, bojnordi2016memristive} architecture.
Since the new memory technology is not very mature, there is no systematic programming method.

The main computational components of these chips usually contain vector-matrix-multiplication units and nonlinear functions, which is the basis of our hardware abstraction.

Moreover, there are quite a few development frameworks for neural networking computing.
Some (like Theano~\cite{2016arXiv160502688full}, TensorFlow~\cite{abadi2016tensorflow}, CNTK~\cite{CNTK}, Torch~\cite{collobert2011torch7}, MXNet~\cite{MXNet}, etc.) describe NN as computation graphs.
For a trained NN, the complete information can be extracted from whichever of them.
\subsection{Neuromorphic chips}\label{sec:neuromorphic_chips}
TrueNorth~\cite{8, 9} is a digital neuromorphic chip for SNNs, based on a structure of tiled crossbar (each crossbar is of size $256 \times 256$, and supports binary-valued neurons and ternary-valued synapses).
The programming paradigm, \emph{Corelet}~\cite{37}, is bound to the hardware platform:
its recent study~\cite{12} proposes a TrueNorth-specific training mechanism to construct the whole NN from top to bottom.
The parameters learned are then mapped to hardware~\cite{38} using \emph{Corelet}.

EMBRACE~\cite{10} is a compact hardware SNN architecture, with the limited fan-out of individual neuron circuits.
From the programming aspect, it follows the Modular Neural Network (MNN) computing paradigm~\cite{39}.
Thus, it is not a general solution.
Neurogrid~\cite{40} and FACETS~\cite{42} (including its successor BrainScaleS~\cite{43}) are analog/digital hybrid systems, whose development methods are both hardware-specific.
SpiNNaker~\cite{44} is different: its toolchain is not bound to any computing paradigm.
The reason is that it is based on the chip multiprocessor (CMP) of ARM cores.
Thus, its neural computing is completed by software.
The drawback is that the efficiency will be lower than the dedicated hardware.
TianJi~\cite{11} is an experimental CMOS neuromorphic chip based on tiled crossbars and supports hybrid computing of ANN and SNN.
A toolchain NEUTRAMS~\cite{NEUTRAMS} is developed to map various NN models onto the chip.
It also decouples application from hardware and completes SW/HW co-design for optimization.
But its methodology is based on redundancy of NN models and is not suitable for large-scale network under strict constraints.

\section{Conclusion}\label{sec:conclusion}
We present a programming solution for NN chips, which can transform a trained, unrestricted NN into an equivalent network to meet hardware constraints.
Multiple techniques are proposed to reduce the transformation error and improve the processing speed.
The solution is validated on a real neuromorphic chip and a PIM design for ANNs, as well as on different scales of NNs under different constraints.
The evaluation shows that the transformation methodology is very effective and only insignificant errors will be introduced. 
The transformation time is much faster than re-training the NN models for a specific neuromorphic hardware.

\begin{acks}                            


The work is supported by the \grantsponsor{GS100000001}{National Key Research and Development Program of China}{http://www.most.gov.cn/eng/programmes1/200610/t20061009_36224.htm} under Grant No.~\grantnum{GS100000001}{2016YFB0200505} and by \grantsponsor{GS100000002}{Beijing Municipal Commission of Science and Technology}{http://www.ebeijing.gov.cn/Government/Departments/t930030.htm} under Grant No.~\grantnum{GS100000002}{Z161100000216147}.

\end{acks}

\balance
\bibliography{references}


\begin{thebibliography}{68}


\ifx \showCODEN    \undefined \def \showCODEN     #1{\unskip}     \fi
\ifx \showDOI      \undefined \def \showDOI       #1{#1}\fi
\ifx \showISBNx    \undefined \def \showISBNx     #1{\unskip}     \fi
\ifx \showISBNxiii \undefined \def \showISBNxiii  #1{\unskip}     \fi
\ifx \showISSN     \undefined \def \showISSN      #1{\unskip}     \fi
\ifx \showLCCN     \undefined \def \showLCCN      #1{\unskip}     \fi
\ifx \shownote     \undefined \def \shownote      #1{#1}          \fi
\ifx \showarticletitle \undefined \def \showarticletitle #1{#1}   \fi
\ifx \showURL      \undefined \def \showURL       {\relax}        \fi
\providecommand\bibfield[2]{#2}
\providecommand\bibinfo[2]{#2}
\providecommand\natexlab[1]{#1}
\providecommand\showeprint[2][]{arXiv:#2}

\bibitem[\protect\citeauthoryear{Abadi, Agarwal, Barham, Brevdo, Chen, Citro,
  Corrado, Davis, Dean, Devin, Ghemawat, Goodfellow, Harp, Irving, Isard, Jia,
  J{\'{o}}zefowicz, Kaiser, Kudlur, Levenberg, Man{\'{e}}, Monga, Moore,
  Murray, Olah, Schuster, Shlens, Steiner, Sutskever, Talwar, Tucker,
  Vanhoucke, Vasudevan, Vi{\'{e}}gas, Vinyals, Warden, Wattenberg, Wicke, Yu,
  and Zheng}{Abadi et~al\mbox{.}}{2016}]%
        {abadi2016tensorflow}
\bibfield{author}{\bibinfo{person}{Mart{\'{\i}}n Abadi},
  \bibinfo{person}{Ashish Agarwal}, \bibinfo{person}{Paul Barham},
  \bibinfo{person}{Eugene Brevdo}, \bibinfo{person}{Zhifeng Chen},
  \bibinfo{person}{Craig Citro}, \bibinfo{person}{Gregory~S. Corrado},
  \bibinfo{person}{Andy Davis}, \bibinfo{person}{Jeffrey Dean},
  \bibinfo{person}{Matthieu Devin}, \bibinfo{person}{Sanjay Ghemawat},
  \bibinfo{person}{Ian~J. Goodfellow}, \bibinfo{person}{Andrew Harp},
  \bibinfo{person}{Geoffrey Irving}, \bibinfo{person}{Michael Isard},
  \bibinfo{person}{Yangqing Jia}, \bibinfo{person}{Rafal J{\'{o}}zefowicz},
  \bibinfo{person}{Lukasz Kaiser}, \bibinfo{person}{Manjunath Kudlur},
  \bibinfo{person}{Josh Levenberg}, \bibinfo{person}{Dan Man{\'{e}}},
  \bibinfo{person}{Rajat Monga}, \bibinfo{person}{Sherry Moore},
  \bibinfo{person}{Derek~Gordon Murray}, \bibinfo{person}{Chris Olah},
  \bibinfo{person}{Mike Schuster}, \bibinfo{person}{Jonathon Shlens},
  \bibinfo{person}{Benoit Steiner}, \bibinfo{person}{Ilya Sutskever},
  \bibinfo{person}{Kunal Talwar}, \bibinfo{person}{Paul~A. Tucker},
  \bibinfo{person}{Vincent Vanhoucke}, \bibinfo{person}{Vijay Vasudevan},
  \bibinfo{person}{Fernanda~B. Vi{\'{e}}gas}, \bibinfo{person}{Oriol Vinyals},
  \bibinfo{person}{Pete Warden}, \bibinfo{person}{Martin Wattenberg},
  \bibinfo{person}{Martin Wicke}, \bibinfo{person}{Yuan Yu}, {and}
  \bibinfo{person}{Xiaoqiang Zheng}.} \bibinfo{year}{2016}\natexlab{}.
\newblock \showarticletitle{TensorFlow: Large-Scale Machine Learning on
  Heterogeneous Distributed Systems}.
\newblock \bibinfo{journal}{\emph{CoRR}}  \bibinfo{volume}{abs/1603.04467}
  (\bibinfo{year}{2016}).
\newblock
\urldef\tempurl%
\url{http://arxiv.org/abs/1603.04467}
\showURL{%
\tempurl}


\bibitem[\protect\citeauthoryear{Agarwal, Akchurin, and Basoglu}{Agarwal
  et~al\mbox{.}}{2014}]%
        {CNTK}
\bibfield{author}{\bibinfo{person}{A. Agarwal}, \bibinfo{person}{E. Akchurin},
  {and} \bibinfo{person}{C. Basoglu}.} \bibinfo{year}{2014}\natexlab{}.
\newblock \showarticletitle{An introduction to computational networks and the
  computational network toolkit}.
\newblock  (\bibinfo{year}{2014}).
\newblock


\bibitem[\protect\citeauthoryear{Akopyan, Sawada, Cassidy, Alvarez-Icaza,
  Arthur, Merolla, Imam, Nakamura, Datta, Nam, Taba, Beakes, Brezzo, Kuang,
  Manohar, Risk, Jackson, and Modha}{Akopyan et~al\mbox{.}}{2015}]%
        {38}
\bibfield{author}{\bibinfo{person}{Filipp Akopyan}, \bibinfo{person}{Jun
  Sawada}, \bibinfo{person}{Andrew Cassidy}, \bibinfo{person}{Rodrigo
  Alvarez-Icaza}, \bibinfo{person}{John Arthur}, \bibinfo{person}{Paul
  Merolla}, \bibinfo{person}{Nabil Imam}, \bibinfo{person}{Yutaka Nakamura},
  \bibinfo{person}{Pallab Datta}, \bibinfo{person}{Gi-Joon Nam},
  \bibinfo{person}{Brian Taba}, \bibinfo{person}{Michael Beakes},
  \bibinfo{person}{Bernard Brezzo}, \bibinfo{person}{Jente~B Kuang},
  \bibinfo{person}{Rajit Manohar}, \bibinfo{person}{William~P Risk},
  \bibinfo{person}{Bryan Jackson}, {and} \bibinfo{person}{Dharmendra~S Modha}.}
  \bibinfo{year}{2015}\natexlab{}.
\newblock \showarticletitle{Truenorth: Design and tool flow of a 65 mw 1
  million neuron programmable neurosynaptic chip}.
\newblock \bibinfo{journal}{\emph{IEEE Transactions on Computer-Aided Design of
  Integrated Circuits and Systems}} \bibinfo{volume}{34}, \bibinfo{number}{10}
  (\bibinfo{year}{2015}), \bibinfo{pages}{1537--1557}.
\newblock


\bibitem[\protect\citeauthoryear{Al{-}Rfou, Alain, Almahairi,
  Angerm{\"{u}}ller, Bahdanau, Ballas, Bastien, Bayer, Belikov, Belopolsky,
  Bengio, Bergeron, Bergstra, Bisson, Snyder, Bouchard,
  Boulanger{-}Lewandowski, Bouthillier, de~Br{\'{e}}bisson, Breuleux, Carrier,
  Cho, Chorowski, Christiano, Cooijmans, C{\^{o}}t{\'{e}}, C{\^{o}}t{\'{e}},
  Courville, Dauphin, Delalleau, Demouth, Desjardins, Dieleman, Dinh, Ducoffe,
  Dumoulin, Kahou, Erhan, Fan, Firat, Germain, Glorot, Goodfellow, Graham,
  G{\"{u}}l{\c{c}}ehre, Hamel, Harlouchet, Heng, Hidasi, Honari, Jain, Jean,
  Jia, Korobov, Kulkarni, Lamb, Lamblin, Larsen, Laurent, Lee,
  Lefran{\c{c}}ois, Lemieux, L{\'{e}}onard, Lin, Livezey, Lorenz, Lowin, Ma,
  Manzagol, Mastropietro, McGibbon, Memisevic, van Merri{\"{e}}nboer,
  Michalski, Mirza, Orlandi, Pal, Pascanu, Pezeshki, Raffel, Renshaw, Rocklin,
  Romero, Roth, Sadowski, Salvatier, Savard, Schl{\"{u}}ter, Schulman,
  Schwartz, Serban, Serdyuk, Shabanian, Simon, Spieckermann, Subramanyam,
  Sygnowski, Tanguay, van Tulder, Turian, Urban, Vincent, Visin, de~Vries,
  Warde{-}Farley, Webb, Willson, Xu, Xue, Yao, Zhang, and Zhang}{Al{-}Rfou
  et~al\mbox{.}}{2016}]%
        {2016arXiv160502688full}
\bibfield{author}{\bibinfo{person}{Rami Al{-}Rfou}, \bibinfo{person}{Guillaume
  Alain}, \bibinfo{person}{Amjad Almahairi}, \bibinfo{person}{Christof
  Angerm{\"{u}}ller}, \bibinfo{person}{Dzmitry Bahdanau},
  \bibinfo{person}{Nicolas Ballas}, \bibinfo{person}{Fr{\'{e}}d{\'{e}}ric
  Bastien}, \bibinfo{person}{Justin Bayer}, \bibinfo{person}{Anatoly Belikov},
  \bibinfo{person}{Alexander Belopolsky}, \bibinfo{person}{Yoshua Bengio},
  \bibinfo{person}{Arnaud Bergeron}, \bibinfo{person}{James Bergstra},
  \bibinfo{person}{Valentin Bisson}, \bibinfo{person}{Josh~Bleecher Snyder},
  \bibinfo{person}{Nicolas Bouchard}, \bibinfo{person}{Nicolas
  Boulanger{-}Lewandowski}, \bibinfo{person}{Xavier Bouthillier},
  \bibinfo{person}{Alexandre de Br{\'{e}}bisson}, \bibinfo{person}{Olivier
  Breuleux}, \bibinfo{person}{Pierre~Luc Carrier}, \bibinfo{person}{Kyunghyun
  Cho}, \bibinfo{person}{Jan Chorowski}, \bibinfo{person}{Paul Christiano},
  \bibinfo{person}{Tim Cooijmans}, \bibinfo{person}{Marc{-}Alexandre
  C{\^{o}}t{\'{e}}}, \bibinfo{person}{Myriam C{\^{o}}t{\'{e}}},
  \bibinfo{person}{Aaron~C. Courville}, \bibinfo{person}{Yann~N. Dauphin},
  \bibinfo{person}{Olivier Delalleau}, \bibinfo{person}{Julien Demouth},
  \bibinfo{person}{Guillaume Desjardins}, \bibinfo{person}{Sander Dieleman},
  \bibinfo{person}{Laurent Dinh}, \bibinfo{person}{Melanie Ducoffe},
  \bibinfo{person}{Vincent Dumoulin}, \bibinfo{person}{Samira~Ebrahimi Kahou},
  \bibinfo{person}{Dumitru Erhan}, \bibinfo{person}{Ziye Fan},
  \bibinfo{person}{Orhan Firat}, \bibinfo{person}{Mathieu Germain},
  \bibinfo{person}{Xavier Glorot}, \bibinfo{person}{Ian~J. Goodfellow},
  \bibinfo{person}{Matthew Graham}, \bibinfo{person}{{\c{C}}aglar
  G{\"{u}}l{\c{c}}ehre}, \bibinfo{person}{Philippe Hamel},
  \bibinfo{person}{Iban Harlouchet}, \bibinfo{person}{Jean{-}Philippe Heng},
  \bibinfo{person}{Bal{\'{a}}zs Hidasi}, \bibinfo{person}{Sina Honari},
  \bibinfo{person}{Arjun Jain}, \bibinfo{person}{S{\'{e}}bastien Jean},
  \bibinfo{person}{Kai Jia}, \bibinfo{person}{Mikhail Korobov},
  \bibinfo{person}{Vivek Kulkarni}, \bibinfo{person}{Alex Lamb},
  \bibinfo{person}{Pascal Lamblin}, \bibinfo{person}{Eric Larsen},
  \bibinfo{person}{C{\'{e}}sar Laurent}, \bibinfo{person}{Sean Lee},
  \bibinfo{person}{Simon Lefran{\c{c}}ois}, \bibinfo{person}{Simon Lemieux},
  \bibinfo{person}{Nicholas L{\'{e}}onard}, \bibinfo{person}{Zhouhan Lin},
  \bibinfo{person}{Jesse~A. Livezey}, \bibinfo{person}{Cory Lorenz},
  \bibinfo{person}{Jeremiah Lowin}, \bibinfo{person}{Qianli Ma},
  \bibinfo{person}{Pierre{-}Antoine Manzagol}, \bibinfo{person}{Olivier
  Mastropietro}, \bibinfo{person}{Robert McGibbon}, \bibinfo{person}{Roland
  Memisevic}, \bibinfo{person}{Bart van Merri{\"{e}}nboer},
  \bibinfo{person}{Vincent Michalski}, \bibinfo{person}{Mehdi Mirza},
  \bibinfo{person}{Alberto Orlandi}, \bibinfo{person}{Christopher~Joseph Pal},
  \bibinfo{person}{Razvan Pascanu}, \bibinfo{person}{Mohammad Pezeshki},
  \bibinfo{person}{Colin Raffel}, \bibinfo{person}{Daniel Renshaw},
  \bibinfo{person}{Matthew Rocklin}, \bibinfo{person}{Adriana Romero},
  \bibinfo{person}{Markus Roth}, \bibinfo{person}{Peter Sadowski},
  \bibinfo{person}{John Salvatier}, \bibinfo{person}{Fran{\c{c}}ois Savard},
  \bibinfo{person}{Jan Schl{\"{u}}ter}, \bibinfo{person}{John Schulman},
  \bibinfo{person}{Gabriel Schwartz}, \bibinfo{person}{Iulian~Vlad Serban},
  \bibinfo{person}{Dmitriy Serdyuk}, \bibinfo{person}{Samira Shabanian},
  \bibinfo{person}{{\'{E}}tienne Simon}, \bibinfo{person}{Sigurd Spieckermann},
  \bibinfo{person}{S.~Ramana Subramanyam}, \bibinfo{person}{Jakub Sygnowski},
  \bibinfo{person}{J{\'{e}}r{\'{e}}mie Tanguay}, \bibinfo{person}{Gijs van
  Tulder}, \bibinfo{person}{Joseph~P. Turian}, \bibinfo{person}{Sebastian
  Urban}, \bibinfo{person}{Pascal Vincent}, \bibinfo{person}{Francesco Visin},
  \bibinfo{person}{Harm de Vries}, \bibinfo{person}{David Warde{-}Farley},
  \bibinfo{person}{Dustin~J. Webb}, \bibinfo{person}{Matthew Willson},
  \bibinfo{person}{Kelvin Xu}, \bibinfo{person}{Lijun Xue}, \bibinfo{person}{Li
  Yao}, \bibinfo{person}{Saizheng Zhang}, {and} \bibinfo{person}{Ying Zhang}.}
  \bibinfo{year}{2016}\natexlab{}.
\newblock \showarticletitle{Theano: {A} Python framework for fast computation
  of mathematical expressions}.
\newblock \bibinfo{journal}{\emph{CoRR}}  \bibinfo{volume}{abs/1605.02688}
  (\bibinfo{year}{2016}).
\newblock
\urldef\tempurl%
\url{http://arxiv.org/abs/1605.02688}
\showURL{%
\tempurl}


\bibitem[\protect\citeauthoryear{Amir, Datta, Risk, Cassidy, Kusnitz, Esser,
  Andreopoulos, Wong, Flickner, Alvarez-Icaza, McQuinn, Shaw, Pass, and
  Modha}{Amir et~al\mbox{.}}{2013}]%
        {37}
\bibfield{author}{\bibinfo{person}{Arnon Amir}, \bibinfo{person}{Pallab Datta},
  \bibinfo{person}{William~P Risk}, \bibinfo{person}{Andrew~S Cassidy},
  \bibinfo{person}{Jeffrey~A Kusnitz}, \bibinfo{person}{Steve~K Esser},
  \bibinfo{person}{Alexander Andreopoulos}, \bibinfo{person}{Theodore~M Wong},
  \bibinfo{person}{Myron Flickner}, \bibinfo{person}{Rodrigo Alvarez-Icaza},
  \bibinfo{person}{Emmett McQuinn}, \bibinfo{person}{Ben Shaw},
  \bibinfo{person}{Norm Pass}, {and} \bibinfo{person}{Dharmendra~S Modha}.}
  \bibinfo{year}{2013}\natexlab{}.
\newblock \showarticletitle{Cognitive computing programming paradigm: a corelet
  language for composing networks of neurosynaptic cores}. In
  \bibinfo{booktitle}{\emph{Neural Networks (IJCNN), The 2013 International
  Joint Conference on}}. IEEE, \bibinfo{pages}{1--10}.
\newblock


\bibitem[\protect\citeauthoryear{Anwar, Hwang, and Sung}{Anwar
  et~al\mbox{.}}{2015}]%
        {anwar2015fixed}
\bibfield{author}{\bibinfo{person}{Sajid Anwar}, \bibinfo{person}{Kyuyeon
  Hwang}, {and} \bibinfo{person}{Wonyong Sung}.}
  \bibinfo{year}{2015}\natexlab{}.
\newblock \showarticletitle{Fixed point optimization of deep convolutional
  neural networks for object recognition}. In
  \bibinfo{booktitle}{\emph{Acoustics, Speech and Signal Processing (ICASSP),
  2015 IEEE International Conference on}}. IEEE, \bibinfo{pages}{1131--1135}.
\newblock


\bibitem[\protect\citeauthoryear{Benjamin, Gao, McQuinn, Choudhary,
  Chandrasekaran, Bussat, Alvarez-Icaza, Arthur, Merolla, and Boahen}{Benjamin
  et~al\mbox{.}}{2014}]%
        {40}
\bibfield{author}{\bibinfo{person}{Ben~Varkey Benjamin},
  \bibinfo{person}{Peiran Gao}, \bibinfo{person}{Emmett McQuinn},
  \bibinfo{person}{Swadesh Choudhary}, \bibinfo{person}{Anand~R
  Chandrasekaran}, \bibinfo{person}{Jean-Marie Bussat},
  \bibinfo{person}{Rodrigo Alvarez-Icaza}, \bibinfo{person}{John~V Arthur},
  \bibinfo{person}{Paul~A Merolla}, {and} \bibinfo{person}{Kwabena Boahen}.}
  \bibinfo{year}{2014}\natexlab{}.
\newblock \showarticletitle{Neurogrid: A mixed-analog-digital multichip system
  for large-scale neural simulations}.
\newblock \bibinfo{journal}{\emph{Proc. IEEE}} \bibinfo{volume}{102},
  \bibinfo{number}{5} (\bibinfo{year}{2014}), \bibinfo{pages}{699--716}.
\newblock


\bibitem[\protect\citeauthoryear{Bojnordi and Ipek}{Bojnordi and Ipek}{2016}]%
        {bojnordi2016memristive}
\bibfield{author}{\bibinfo{person}{Mahdi~Nazm Bojnordi} {and}
  \bibinfo{person}{Engin Ipek}.} \bibinfo{year}{2016}\natexlab{}.
\newblock \showarticletitle{Memristive Boltzmann machine: A hardware
  accelerator for combinatorial optimization and deep learning}. In
  \bibinfo{booktitle}{\emph{High Performance Computer Architecture (HPCA), 2016
  IEEE International Symposium on}}. \bibinfo{pages}{1--13}.
\newblock


\bibitem[\protect\citeauthoryear{Carrillo, Harkin, McDaid, Morgan, Pande,
  Cawley, and McGinley}{Carrillo et~al\mbox{.}}{2013}]%
        {10}
\bibfield{author}{\bibinfo{person}{Snaider Carrillo}, \bibinfo{person}{Jim
  Harkin}, \bibinfo{person}{Liam~J McDaid}, \bibinfo{person}{Fearghal Morgan},
  \bibinfo{person}{Sandeep Pande}, \bibinfo{person}{Seamus Cawley}, {and}
  \bibinfo{person}{Brian McGinley}.} \bibinfo{year}{2013}\natexlab{}.
\newblock \showarticletitle{Scalable hierarchical network-on-chip architecture
  for spiking neural network hardware implementations}.
\newblock \bibinfo{journal}{\emph{IEEE Transactions on Parallel and Distributed
  Systems}} \bibinfo{volume}{24}, \bibinfo{number}{12} (\bibinfo{year}{2013}),
  \bibinfo{pages}{2451--2461}.
\newblock


\bibitem[\protect\citeauthoryear{Cassidy, Merolla, Arthur, Esser, Jackson,
  Alvarez-Icaza, Datta, Sawada, Wong, Feldman, Amir, Rubin, Akopyan, McQuinn,
  Risk, and Modha}{Cassidy et~al\mbox{.}}{2013}]%
        {9}
\bibfield{author}{\bibinfo{person}{Andrew~S Cassidy}, \bibinfo{person}{Paul
  Merolla}, \bibinfo{person}{John~V Arthur}, \bibinfo{person}{Steve~K Esser},
  \bibinfo{person}{Bryan Jackson}, \bibinfo{person}{Rodrigo Alvarez-Icaza},
  \bibinfo{person}{Pallab Datta}, \bibinfo{person}{Jun Sawada},
  \bibinfo{person}{Theodore~M Wong}, \bibinfo{person}{Vitaly Feldman},
  \bibinfo{person}{Arnon Amir}, \bibinfo{person}{Daniel Ben-Dayan Rubin},
  \bibinfo{person}{Filipp Akopyan}, \bibinfo{person}{Emmett McQuinn},
  \bibinfo{person}{William~P Risk}, {and} \bibinfo{person}{Dharmendra~S
  Modha}.} \bibinfo{year}{2013}\natexlab{}.
\newblock \showarticletitle{Cognitive computing building block: A versatile and
  efficient digital neuron model for neurosynaptic cores}. In
  \bibinfo{booktitle}{\emph{Neural Networks (IJCNN), The 2013 International
  Joint Conference on}}. IEEE, \bibinfo{pages}{1--10}.
\newblock


\bibitem[\protect\citeauthoryear{Cavigelli and Benini}{Cavigelli and
  Benini}{2016}]%
        {22}
\bibfield{author}{\bibinfo{person}{Lukas Cavigelli} {and} \bibinfo{person}{Luca
  Benini}.} \bibinfo{year}{2016}\natexlab{}.
\newblock \showarticletitle{A 803 gop/s/w convolutional network accelerator}.
\newblock \bibinfo{journal}{\emph{IEEE Transactions on Circuits and Systems for
  Video Technology}} (\bibinfo{year}{2016}).
\newblock


\bibitem[\protect\citeauthoryear{Chen, Du, Sun, Wang, Wu, Chen, and Temam}{Chen
  et~al\mbox{.}}{2014a}]%
        {13}
\bibfield{author}{\bibinfo{person}{Tianshi Chen}, \bibinfo{person}{Zidong Du},
  \bibinfo{person}{Ninghui Sun}, \bibinfo{person}{Jia Wang},
  \bibinfo{person}{Chengyong Wu}, \bibinfo{person}{Yunji Chen}, {and}
  \bibinfo{person}{Olivier Temam}.} \bibinfo{year}{2014}\natexlab{a}.
\newblock \showarticletitle{Diannao: A small-footprint high-throughput
  accelerator for ubiquitous machine-learning}. In
  \bibinfo{booktitle}{\emph{ACM Sigplan Notices}}, Vol.~\bibinfo{volume}{49}.
  ACM, \bibinfo{pages}{269--284}.
\newblock


\bibitem[\protect\citeauthoryear{Chen, Li, Li, Lin, Wang, Wang, Xiao, Xu,
  Zhang, and Zhang}{Chen et~al\mbox{.}}{2015b}]%
        {MXNet}
\bibfield{author}{\bibinfo{person}{Tianqi Chen}, \bibinfo{person}{Mu Li},
  \bibinfo{person}{Yutian Li}, \bibinfo{person}{Min Lin},
  \bibinfo{person}{Naiyan Wang}, \bibinfo{person}{Minjie Wang},
  \bibinfo{person}{Tianjun Xiao}, \bibinfo{person}{Bing Xu},
  \bibinfo{person}{Chiyuan Zhang}, {and} \bibinfo{person}{Zheng Zhang}.}
  \bibinfo{year}{2015}\natexlab{b}.
\newblock \showarticletitle{Mxnet: A flexible and efficient machine learning
  library for heterogeneous distributed systems}.
\newblock \bibinfo{journal}{\emph{arXiv preprint arXiv:1512.01274}}.
\newblock


\bibitem[\protect\citeauthoryear{Chen, Wilson, Tyree, Weinberger, and
  Chen}{Chen et~al\mbox{.}}{2015c}]%
        {chen2015compressing}
\bibfield{author}{\bibinfo{person}{Wenlin Chen}, \bibinfo{person}{James
  Wilson}, \bibinfo{person}{Stephen Tyree}, \bibinfo{person}{Kilian
  Weinberger}, {and} \bibinfo{person}{Yixin Chen}.}
  \bibinfo{year}{2015}\natexlab{c}.
\newblock \showarticletitle{Compressing neural networks with the hashing
  trick}. In \bibinfo{booktitle}{\emph{International Conference on Machine
  Learning}}. \bibinfo{pages}{2285--2294}.
\newblock


\bibitem[\protect\citeauthoryear{Chen, Luo, Liu, Zhang, He, Wang, Li, Chen, Xu,
  Sun, and Teman}{Chen et~al\mbox{.}}{2014b}]%
        {15}
\bibfield{author}{\bibinfo{person}{Yunji Chen}, \bibinfo{person}{Tao Luo},
  \bibinfo{person}{Shaoli Liu}, \bibinfo{person}{Shijin Zhang},
  \bibinfo{person}{Liqiang He}, \bibinfo{person}{Jia Wang},
  \bibinfo{person}{Ling Li}, \bibinfo{person}{Tianshi Chen},
  \bibinfo{person}{Zhiwei Xu}, \bibinfo{person}{Ninghui Sun}, {and}
  \bibinfo{person}{Olivier Teman}.} \bibinfo{year}{2014}\natexlab{b}.
\newblock \showarticletitle{Dadiannao: A machine-learning supercomputer}. In
  \bibinfo{booktitle}{\emph{Proceedings of the 47th Annual IEEE/ACM
  International Symposium on Microarchitecture}}. IEEE Computer Society,
  \bibinfo{pages}{609--622}.
\newblock


\bibitem[\protect\citeauthoryear{Chen, Krishna, Emer, and Sze}{Chen
  et~al\mbox{.}}{2016}]%
        {16}
\bibfield{author}{\bibinfo{person}{Y.~H. Chen}, \bibinfo{person}{T. Krishna},
  \bibinfo{person}{J. Emer}, {and} \bibinfo{person}{V. Sze}.}
  \bibinfo{year}{2016}\natexlab{}.
\newblock \showarticletitle{14.5 Eyeriss: An energy-efficient reconfigurable
  accelerator for deep convolutional neural networks}. In
  \bibinfo{booktitle}{\emph{2016 IEEE International Solid-State Circuits
  Conference (ISSCC)}}. \bibinfo{pages}{262--263}.
\newblock
\urldef\tempurl%
\url{https://doi.org/10.1109/ISSCC.2016.7418007}
\showDOI{\tempurl}


\bibitem[\protect\citeauthoryear{Chen, Gao, Zhou, Huang, Li, and Ma}{Chen
  et~al\mbox{.}}{2015a}]%
        {chen2015optimized}
\bibfield{author}{\bibinfo{person}{Zhen Chen}, \bibinfo{person}{Bin Gao},
  \bibinfo{person}{Zheng Zhou}, \bibinfo{person}{Peng Huang},
  \bibinfo{person}{Haitong Li}, {and} \bibinfo{person}{Wenjia Ma}.}
  \bibinfo{year}{2015}\natexlab{a}.
\newblock \showarticletitle{Optimized learning scheme for grayscale image
  recognition in a RRAM based analog neuromorphic system}. In
  \bibinfo{booktitle}{\emph{Electron Devices Meeting (IEDM), 2015 IEEE
  International}}. IEEE.
\newblock


\bibitem[\protect\citeauthoryear{Chi, Li, Xu, Zhang, Zhao, Liu, Wang, and
  Xie}{Chi et~al\mbox{.}}{2016}]%
        {7}
\bibfield{author}{\bibinfo{person}{Ping Chi}, \bibinfo{person}{Shuangchen Li},
  \bibinfo{person}{Cong Xu}, \bibinfo{person}{Tao Zhang},
  \bibinfo{person}{Jishen Zhao}, \bibinfo{person}{Yongpan Liu},
  \bibinfo{person}{Yu Wang}, {and} \bibinfo{person}{Yuan Xie}.}
  \bibinfo{year}{2016}\natexlab{}.
\newblock \showarticletitle{Prime: A novel processing-in-memory architecture
  for neural network computation in reram-based main memory}. In
  \bibinfo{booktitle}{\emph{Proceedings of the 43rd International Symposium on
  Computer Architecture}}. IEEE Press, \bibinfo{pages}{27--39}.
\newblock


\bibitem[\protect\citeauthoryear{Collobert, Kavukcuoglu, and Farabet}{Collobert
  et~al\mbox{.}}{2011}]%
        {collobert2011torch7}
\bibfield{author}{\bibinfo{person}{Ronan Collobert}, \bibinfo{person}{Koray
  Kavukcuoglu}, {and} \bibinfo{person}{Clement Farabet}.}
  \bibinfo{year}{2011}\natexlab{}.
\newblock \showarticletitle{Torch7: A Matlab-like Environment for Machine
  Learning}. In \bibinfo{booktitle}{\emph{neural information processing
  systems}}.
\newblock


\bibitem[\protect\citeauthoryear{Denil, Shakibi, Dinh, Ranzato, and
  de~Freitas}{Denil et~al\mbox{.}}{2013}]%
        {denil2013predicting}
\bibfield{author}{\bibinfo{person}{Misha Denil}, \bibinfo{person}{Babak
  Shakibi}, \bibinfo{person}{Laurent Dinh},
  \bibinfo{person}{Marc\textquotesingle~Aurelio Ranzato}, {and}
  \bibinfo{person}{Nando de Freitas}.} \bibinfo{year}{2013}\natexlab{}.
\newblock \showarticletitle{Predicting Parameters in Deep Learning}. In
  \bibinfo{booktitle}{\emph{Advances in Neural Information Processing Systems
  26}}, \bibfield{editor}{\bibinfo{person}{C.~J.~C. Burges},
  \bibinfo{person}{L.~Bottou}, \bibinfo{person}{M.~Welling},
  \bibinfo{person}{Z.~Ghahramani}, {and} \bibinfo{person}{K.~Q. Weinberger}}
  (Eds.). \bibinfo{publisher}{Curran Associates, Inc.},
  \bibinfo{pages}{2148--2156}.
\newblock


\bibitem[\protect\citeauthoryear{Du, Fasthuber, Chen, Ienne, Li, Luo, Feng,
  Chen, and Temam}{Du et~al\mbox{.}}{2015}]%
        {21}
\bibfield{author}{\bibinfo{person}{Zidong Du}, \bibinfo{person}{Robert
  Fasthuber}, \bibinfo{person}{Tianshi Chen}, \bibinfo{person}{Paolo Ienne},
  \bibinfo{person}{Ling Li}, \bibinfo{person}{Tao Luo},
  \bibinfo{person}{Xiaobing Feng}, \bibinfo{person}{Yunji Chen}, {and}
  \bibinfo{person}{Olivier Temam}.} \bibinfo{year}{2015}\natexlab{}.
\newblock \showarticletitle{ShiDianNao: Shifting vision processing closer to
  the sensor}. In \bibinfo{booktitle}{\emph{ACM SIGARCH Computer Architecture
  News}}, Vol.~\bibinfo{volume}{43}. ACM, \bibinfo{pages}{92--104}.
\newblock


\bibitem[\protect\citeauthoryear{Esser, Merolla, Arthur, Cassidy, Appuswamy,
  Andreopoulos, Berg, McKinstry, Melano, Barch, di~Nolfo, Datta, Amir, Taba,
  Flickner, and Modha}{Esser et~al\mbox{.}}{2016}]%
        {12}
\bibfield{author}{\bibinfo{person}{Steven~K Esser}, \bibinfo{person}{Paul~A
  Merolla}, \bibinfo{person}{John~V Arthur}, \bibinfo{person}{Andrew~S
  Cassidy}, \bibinfo{person}{Rathinakumar Appuswamy},
  \bibinfo{person}{Alexander Andreopoulos}, \bibinfo{person}{David~J Berg},
  \bibinfo{person}{Jeffrey~L McKinstry}, \bibinfo{person}{Timothy Melano},
  \bibinfo{person}{Davis~R Barch}, \bibinfo{person}{Carmelo di Nolfo},
  \bibinfo{person}{Pallab Datta}, \bibinfo{person}{Arnon Amir},
  \bibinfo{person}{Brian Taba}, \bibinfo{person}{Myron~D Flickner}, {and}
  \bibinfo{person}{Dharmendra~S Modha}.} \bibinfo{year}{2016}\natexlab{}.
\newblock \showarticletitle{Convolutional networks for fast, energy-efficient
  neuromorphic computing}.
\newblock \bibinfo{journal}{\emph{Proceedings of the National Academy of
  Sciences}} (\bibinfo{year}{2016}), \bibinfo{pages}{201604850}.
\newblock


\bibitem[\protect\citeauthoryear{Farabet, Martini, Corda, Akselrod,
  Culurciello, and LeCun}{Farabet et~al\mbox{.}}{2011}]%
        {31}
\bibfield{author}{\bibinfo{person}{Cl{\'e}ment Farabet}, \bibinfo{person}{Berin
  Martini}, \bibinfo{person}{Benoit Corda}, \bibinfo{person}{Polina Akselrod},
  \bibinfo{person}{Eugenio Culurciello}, {and} \bibinfo{person}{Yann LeCun}.}
  \bibinfo{year}{2011}\natexlab{}.
\newblock \showarticletitle{Neuflow: A runtime reconfigurable dataflow
  processor for vision}. In \bibinfo{booktitle}{\emph{Computer Vision and
  Pattern Recognition Workshops (CVPRW), 2011 IEEE Computer Society Conference
  on}}. IEEE, \bibinfo{pages}{109--116}.
\newblock


\bibitem[\protect\citeauthoryear{Farabet, Poulet, Han, and LeCun}{Farabet
  et~al\mbox{.}}{2009}]%
        {23}
\bibfield{author}{\bibinfo{person}{Cl{\'e}ment Farabet}, \bibinfo{person}{Cyril
  Poulet}, \bibinfo{person}{Jefferson~Y Han}, {and} \bibinfo{person}{Yann
  LeCun}.} \bibinfo{year}{2009}\natexlab{}.
\newblock \showarticletitle{Cnp: An fpga-based processor for convolutional
  networks}. In \bibinfo{booktitle}{\emph{Field Programmable Logic and
  Applications, 2009. FPL 2009. International Conference on}}. IEEE,
  \bibinfo{pages}{32--37}.
\newblock


\bibitem[\protect\citeauthoryear{Furber, Lester, Plana, Garside, Painkras,
  Temple, and Brown}{Furber et~al\mbox{.}}{2013}]%
        {44}
\bibfield{author}{\bibinfo{person}{Steve~B Furber}, \bibinfo{person}{David~R
  Lester}, \bibinfo{person}{Luis~A Plana}, \bibinfo{person}{Jim~D Garside},
  \bibinfo{person}{Eustace Painkras}, \bibinfo{person}{Steve Temple}, {and}
  \bibinfo{person}{Andrew~D Brown}.} \bibinfo{year}{2013}\natexlab{}.
\newblock \showarticletitle{Overview of the spinnaker system architecture}.
\newblock \bibinfo{journal}{\emph{IEEE Trans. Comput.}} \bibinfo{volume}{62},
  \bibinfo{number}{12} (\bibinfo{year}{2013}), \bibinfo{pages}{2454--2467}.
\newblock


\bibitem[\protect\citeauthoryear{Han, Kang, Mao, Hu, Li, Li, Xie, Luo, Yao,
  Wang, Yang, and Dally}{Han et~al\mbox{.}}{2016a}]%
        {han2016ese}
\bibfield{author}{\bibinfo{person}{Song Han}, \bibinfo{person}{Junlong Kang},
  \bibinfo{person}{Huizi Mao}, \bibinfo{person}{Yiming Hu},
  \bibinfo{person}{Xin Li}, \bibinfo{person}{Yubin Li},
  \bibinfo{person}{Dongliang Xie}, \bibinfo{person}{Hong Luo},
  \bibinfo{person}{Song Yao}, \bibinfo{person}{Yu Wang},
  \bibinfo{person}{Huazhong Yang}, {and} \bibinfo{person}{William~J. Dally}.}
  \bibinfo{year}{2016}\natexlab{a}.
\newblock \showarticletitle{{ESE:} Efficient Speech Recognition Engine with
  Compressed {LSTM} on {FPGA}}.
\newblock \bibinfo{journal}{\emph{CoRR}}  \bibinfo{volume}{abs/1612.00694}
  (\bibinfo{year}{2016}).
\newblock
\urldef\tempurl%
\url{http://arxiv.org/abs/1612.00694}
\showURL{%
\tempurl}


\bibitem[\protect\citeauthoryear{Han, Liu, Mao, Pu, Pedram, Horowitz, and
  Dally}{Han et~al\mbox{.}}{2016b}]%
        {han2016eie}
\bibfield{author}{\bibinfo{person}{Song Han}, \bibinfo{person}{Xingyu Liu},
  \bibinfo{person}{Huizi Mao}, \bibinfo{person}{Jing Pu},
  \bibinfo{person}{Ardavan Pedram}, \bibinfo{person}{Mark~A Horowitz}, {and}
  \bibinfo{person}{William~J Dally}.} \bibinfo{year}{2016}\natexlab{b}.
\newblock \showarticletitle{EIE: efficient inference engine on compressed deep
  neural network}. In \bibinfo{booktitle}{\emph{Proceedings of the 43rd
  International Symposium on Computer Architecture}}. IEEE Press,
  \bibinfo{pages}{243--254}.
\newblock


\bibitem[\protect\citeauthoryear{Han, Mao, and Dally}{Han
  et~al\mbox{.}}{2015}]%
        {han2015deep}
\bibfield{author}{\bibinfo{person}{Song Han}, \bibinfo{person}{Huizi Mao},
  {and} \bibinfo{person}{William~J Dally}.} \bibinfo{year}{2015}\natexlab{}.
\newblock \showarticletitle{Deep compression: Compressing deep neural networks
  with pruning, trained quantization and huffman coding}.
\newblock \bibinfo{journal}{\emph{arXiv preprint arXiv:1510.00149}}
  (\bibinfo{year}{2015}).
\newblock


\bibitem[\protect\citeauthoryear{Happel and Murre}{Happel and Murre}{1994}]%
        {39}
\bibfield{author}{\bibinfo{person}{Bart~LM Happel} {and}
  \bibinfo{person}{Jacob~MJ Murre}.} \bibinfo{year}{1994}\natexlab{}.
\newblock \showarticletitle{Design and evolution of modular neural network
  architectures}.
\newblock \bibinfo{journal}{\emph{Neural networks}} \bibinfo{volume}{7},
  \bibinfo{number}{6} (\bibinfo{year}{1994}), \bibinfo{pages}{985--1004}.
\newblock


\bibitem[\protect\citeauthoryear{Hornik, Stinchcombe, and White}{Hornik
  et~al\mbox{.}}{1989}]%
        {hornik1989multilayer}
\bibfield{author}{\bibinfo{person}{Kurt Hornik}, \bibinfo{person}{Maxwell
  Stinchcombe}, {and} \bibinfo{person}{Halbert White}.}
  \bibinfo{year}{1989}\natexlab{}.
\newblock \showarticletitle{Multilayer feedforward networks are universal
  approximators}.
\newblock \bibinfo{journal}{\emph{Neural networks}} \bibinfo{volume}{2},
  \bibinfo{number}{5} (\bibinfo{year}{1989}), \bibinfo{pages}{359--366}.
\newblock


\bibitem[\protect\citeauthoryear{Hu, Li, Chen, Wu, and Rose}{Hu
  et~al\mbox{.}}{2013}]%
        {6}
\bibfield{author}{\bibinfo{person}{Miao Hu}, \bibinfo{person}{Hai Li},
  \bibinfo{person}{Yiran Chen}, \bibinfo{person}{Qing Wu}, {and}
  \bibinfo{person}{Garrett~S Rose}.} \bibinfo{year}{2013}\natexlab{}.
\newblock \showarticletitle{BSB training scheme implementation on
  memristor-based circuit}. In \bibinfo{booktitle}{\emph{Computational
  Intelligence for Security and Defense Applications (CISDA), 2013 IEEE
  Symposium on}}. IEEE, \bibinfo{pages}{80--87}.
\newblock


\bibitem[\protect\citeauthoryear{Hu, Strachan, Li, Grafals, Davila, Graves,
  Lam, Ge, Yang, and Williams}{Hu et~al\mbox{.}}{2016}]%
        {hu2016dot}
\bibfield{author}{\bibinfo{person}{Miao Hu}, \bibinfo{person}{John~Paul
  Strachan}, \bibinfo{person}{Zhiyong Li}, \bibinfo{person}{Emmanuelle~M
  Grafals}, \bibinfo{person}{Noraica Davila}, \bibinfo{person}{Catherine
  Graves}, \bibinfo{person}{Sity Lam}, \bibinfo{person}{Ning Ge},
  \bibinfo{person}{Jianhua~Joshua Yang}, {and} \bibinfo{person}{R~Stanley
  Williams}.} \bibinfo{year}{2016}\natexlab{}.
\newblock \showarticletitle{Dot-product engine for neuromorphic computing:
  programming 1T1M crossbar to accelerate matrix-vector multiplication}. In
  \bibinfo{booktitle}{\emph{Design Automation Conference (DAC), 2016 53nd
  ACM/EDAC/IEEE}}. IEEE, \bibinfo{pages}{1--6}.
\newblock


\bibitem[\protect\citeauthoryear{Hunsberger and Eliasmith}{Hunsberger and
  Eliasmith}{2016}]%
        {46}
\bibfield{author}{\bibinfo{person}{Eric Hunsberger} {and}
  \bibinfo{person}{Chris Eliasmith}.} \bibinfo{year}{2016}\natexlab{}.
\newblock \showarticletitle{Training Spiking Deep Networks for Neuromorphic
  Hardware}.
\newblock \bibinfo{journal}{\emph{CoRR}}  \bibinfo{volume}{abs/1611.05141}
  (\bibinfo{year}{2016}).
\newblock
\urldef\tempurl%
\url{http://arxiv.org/abs/1611.05141}
\showURL{%
\tempurl}


\bibitem[\protect\citeauthoryear{Hwang and Sung}{Hwang and Sung}{2014}]%
        {hwang2014fixed}
\bibfield{author}{\bibinfo{person}{Kyuyeon Hwang} {and}
  \bibinfo{person}{Wonyong Sung}.} \bibinfo{year}{2014}\natexlab{}.
\newblock \showarticletitle{Fixed-point feedforward deep neural network design
  using weights+ 1, 0, and- 1}. In \bibinfo{booktitle}{\emph{Signal Processing
  Systems (SiPS), 2014 IEEE Workshop on}}. IEEE, \bibinfo{pages}{1--6}.
\newblock


\bibitem[\protect\citeauthoryear{Ji, Zhang, Li, Chi, Jiang, Qu, Xie, and
  Chen}{Ji et~al\mbox{.}}{2016}]%
        {NEUTRAMS}
\bibfield{author}{\bibinfo{person}{YU Ji}, \bibinfo{person}{Youhui Zhang},
  \bibinfo{person}{ShuangChen Li}, \bibinfo{person}{Ping Chi},
  \bibinfo{person}{CiHang Jiang}, \bibinfo{person}{Peng Qu},
  \bibinfo{person}{Yuan Xie}, {and} \bibinfo{person}{WenGuang Chen}.}
  \bibinfo{year}{2016}\natexlab{}.
\newblock \showarticletitle{NEUTRAMS: Neural Network Transformation and
  Co-design under Neuromorphic Hardware Constraints}. In
  \bibinfo{booktitle}{\emph{Microarchitecture (MICRO), 2016 49th Annual
  IEEE/ACM International Symposium on}}.
\newblock


\bibitem[\protect\citeauthoryear{Jia, Shelhamer, Donahue, Karayev, Long,
  Girshick, Guadarrama, and Darrell}{Jia et~al\mbox{.}}{2014}]%
        {jia2014caffe}
\bibfield{author}{\bibinfo{person}{Yangqing Jia}, \bibinfo{person}{Evan
  Shelhamer}, \bibinfo{person}{Jeff Donahue}, \bibinfo{person}{Sergey Karayev},
  \bibinfo{person}{Jonathan Long}, \bibinfo{person}{Ross Girshick},
  \bibinfo{person}{Sergio Guadarrama}, {and} \bibinfo{person}{Trevor Darrell}.}
  \bibinfo{year}{2014}\natexlab{}.
\newblock \showarticletitle{Caffe: Convolutional architecture for fast feature
  embedding}. In \bibinfo{booktitle}{\emph{Proceedings of the 22nd ACM
  international conference on Multimedia}}. ACM, \bibinfo{pages}{675--678}.
\newblock


\bibitem[\protect\citeauthoryear{Jouppi, Young, Patil, Patterson, Agrawal,
  Bajwa, Bates, Bhatia, Boden, Borchers, Boyle, Cantin, Chao, Clark, Coriell,
  Daley, Dau, Dean, Gelb, Ghaemmaghami, Gottipati, Gulland, Hagmann, Ho,
  Hogberg, Hu, Hundt, Hurt, Ibarz, Jaffey, Jaworski, Kaplan, Khaitan, Koch,
  Kumar, Lacy, Laudon, Law, Le, Leary, Liu, Lucke, Lundin, MacKean, Maggiore,
  Mahony, Miller, Nagarajan, Narayanaswami, Ni, Nix, Norrie, Omernick,
  Penukonda, Phelps, Ross, Salek, Samadiani, Severn, Sizikov, Snelham, Souter,
  Steinberg, Swing, Tan, Thorson, Tian, Toma, Tuttle, Vasudevan, Walter, Wang,
  Wilcox, and Yoon}{Jouppi et~al\mbox{.}}{2017}]%
        {TPU2017}
\bibfield{author}{\bibinfo{person}{Norman~P. Jouppi}, \bibinfo{person}{Cliff
  Young}, \bibinfo{person}{Nishant Patil}, \bibinfo{person}{David Patterson},
  \bibinfo{person}{Gaurav Agrawal}, \bibinfo{person}{Raminder Bajwa},
  \bibinfo{person}{Sarah Bates}, \bibinfo{person}{Suresh Bhatia},
  \bibinfo{person}{Nan Boden}, \bibinfo{person}{Al Borchers},
  \bibinfo{person}{Rick Boyle}, \bibinfo{person}{Pierre{-}luc Cantin},
  \bibinfo{person}{Clifford Chao}, \bibinfo{person}{Chris Clark},
  \bibinfo{person}{Jeremy Coriell}, \bibinfo{person}{Mike Daley},
  \bibinfo{person}{Matt Dau}, \bibinfo{person}{Jeffrey Dean},
  \bibinfo{person}{Ben Gelb}, \bibinfo{person}{Tara~Vazir Ghaemmaghami},
  \bibinfo{person}{Rajendra Gottipati}, \bibinfo{person}{William Gulland},
  \bibinfo{person}{Robert Hagmann}, \bibinfo{person}{Richard~C. Ho},
  \bibinfo{person}{Doug Hogberg}, \bibinfo{person}{John Hu},
  \bibinfo{person}{Robert Hundt}, \bibinfo{person}{Dan Hurt},
  \bibinfo{person}{Julian Ibarz}, \bibinfo{person}{Aaron Jaffey},
  \bibinfo{person}{Alek Jaworski}, \bibinfo{person}{Alexander Kaplan},
  \bibinfo{person}{Harshit Khaitan}, \bibinfo{person}{Andy Koch},
  \bibinfo{person}{Naveen Kumar}, \bibinfo{person}{Steve Lacy},
  \bibinfo{person}{James Laudon}, \bibinfo{person}{James Law},
  \bibinfo{person}{Diemthu Le}, \bibinfo{person}{Chris Leary},
  \bibinfo{person}{Zhuyuan Liu}, \bibinfo{person}{Kyle Lucke},
  \bibinfo{person}{Alan Lundin}, \bibinfo{person}{Gordon MacKean},
  \bibinfo{person}{Adriana Maggiore}, \bibinfo{person}{Maire Mahony},
  \bibinfo{person}{Kieran Miller}, \bibinfo{person}{Rahul Nagarajan},
  \bibinfo{person}{Ravi Narayanaswami}, \bibinfo{person}{Ray Ni},
  \bibinfo{person}{Kathy Nix}, \bibinfo{person}{Thomas Norrie},
  \bibinfo{person}{Mark Omernick}, \bibinfo{person}{Narayana Penukonda},
  \bibinfo{person}{Andy Phelps}, \bibinfo{person}{Jonathan Ross},
  \bibinfo{person}{Amir Salek}, \bibinfo{person}{Emad Samadiani},
  \bibinfo{person}{Chris Severn}, \bibinfo{person}{Gregory Sizikov},
  \bibinfo{person}{Matthew Snelham}, \bibinfo{person}{Jed Souter},
  \bibinfo{person}{Dan Steinberg}, \bibinfo{person}{Andy Swing},
  \bibinfo{person}{Mercedes Tan}, \bibinfo{person}{Gregory Thorson},
  \bibinfo{person}{Bo Tian}, \bibinfo{person}{Horia Toma},
  \bibinfo{person}{Erick Tuttle}, \bibinfo{person}{Vijay Vasudevan},
  \bibinfo{person}{Richard Walter}, \bibinfo{person}{Walter Wang},
  \bibinfo{person}{Eric Wilcox}, {and} \bibinfo{person}{Doe~Hyun Yoon}.}
  \bibinfo{year}{2017}\natexlab{}.
\newblock \showarticletitle{In-Datacenter Performance Analysis of a Tensor
  Processing Unit}.
\newblock \bibinfo{journal}{\emph{CoRR}}  \bibinfo{volume}{abs/1704.04760}.
\newblock
\urldef\tempurl%
\url{http://arxiv.org/abs/1704.04760}
\showURL{%
\tempurl}


\bibitem[\protect\citeauthoryear{Judd, Albericio, Hetherington, Aamodt, and
  Moshovos}{Judd et~al\mbox{.}}{2016}]%
        {judd2016stripes}
\bibfield{author}{\bibinfo{person}{Patrick Judd}, \bibinfo{person}{Jorge
  Albericio}, \bibinfo{person}{Tayler Hetherington}, \bibinfo{person}{Tor~M
  Aamodt}, {and} \bibinfo{person}{Andreas Moshovos}.}
  \bibinfo{year}{2016}\natexlab{}.
\newblock \showarticletitle{Stripes: Bit-serial deep neural network computing}.
  In \bibinfo{booktitle}{\emph{Microarchitecture (MICRO), 2016 49th Annual
  IEEE/ACM International Symposium on}}. IEEE, \bibinfo{pages}{1--12}.
\newblock


\bibitem[\protect\citeauthoryear{Kim, Kung, Chai, Yalamanchili, and
  Mukhopadhyay}{Kim et~al\mbox{.}}{2016}]%
        {29}
\bibfield{author}{\bibinfo{person}{Duckhwan Kim}, \bibinfo{person}{Jaeha Kung},
  \bibinfo{person}{Sek Chai}, \bibinfo{person}{Sudhakar Yalamanchili}, {and}
  \bibinfo{person}{Saibal Mukhopadhyay}.} \bibinfo{year}{2016}\natexlab{}.
\newblock \showarticletitle{Neurocube: A programmable digital neuromorphic
  architecture with high-density 3D memory}. In
  \bibinfo{booktitle}{\emph{Computer Architecture (ISCA), 2016 ACM/IEEE 43rd
  Annual International Symposium on}}. IEEE, \bibinfo{pages}{380--392}.
\newblock


\bibitem[\protect\citeauthoryear{Kim, Zhang, and Li}{Kim et~al\mbox{.}}{2015}]%
        {4}
\bibfield{author}{\bibinfo{person}{Yongtae Kim}, \bibinfo{person}{Yong Zhang},
  {and} \bibinfo{person}{Peng Li}.} \bibinfo{year}{2015}\natexlab{}.
\newblock \showarticletitle{A Reconfigurable Digital Neuromorphic Processor
  with Memristive Synaptic Crossbar for Cognitive Computing}.
\newblock \bibinfo{journal}{\emph{J. Emerg. Technol. Comput. Syst.}}
  \bibinfo{volume}{11}, \bibinfo{number}{4}, Article \bibinfo{articleno}{38}
  (\bibinfo{date}{April} \bibinfo{year}{2015}), \bibinfo{numpages}{25}~pages.
\newblock
\showISSN{1550-4832}
\urldef\tempurl%
\url{https://doi.org/10.1145/2700234}
\showDOI{\tempurl}


\bibitem[\protect\citeauthoryear{Krizhevsky, Sutskever, and Hinton}{Krizhevsky
  et~al\mbox{.}}{2012}]%
        {AlexNet}
\bibfield{author}{\bibinfo{person}{Alex Krizhevsky}, \bibinfo{person}{Ilya
  Sutskever}, {and} \bibinfo{person}{Geoffrey~E Hinton}.}
  \bibinfo{year}{2012}\natexlab{}.
\newblock \showarticletitle{ImageNet Classification with Deep Convolutional
  Neural Networks}.
\newblock In \bibinfo{booktitle}{\emph{Advances in Neural Information
  Processing Systems 25}}, \bibfield{editor}{\bibinfo{person}{F.~Pereira},
  \bibinfo{person}{C.~J.~C. Burges}, \bibinfo{person}{L.~Bottou}, {and}
  \bibinfo{person}{K.~Q. Weinberger}} (Eds.). \bibinfo{publisher}{Curran
  Associates, Inc.}, \bibinfo{pages}{1097--1105}.
\newblock


\bibitem[\protect\citeauthoryear{LeCun, Bottou, Bengio, and Haffner}{LeCun
  et~al\mbox{.}}{1998}]%
        {LeNet}
\bibfield{author}{\bibinfo{person}{Yann LeCun}, \bibinfo{person}{L{\'e}on
  Bottou}, \bibinfo{person}{Yoshua Bengio}, {and} \bibinfo{person}{Patrick
  Haffner}.} \bibinfo{year}{1998}\natexlab{}.
\newblock \showarticletitle{Gradient-based learning applied to document
  recognition}.
\newblock \bibinfo{journal}{\emph{Proc. IEEE}} \bibinfo{volume}{86},
  \bibinfo{number}{11} (\bibinfo{year}{1998}), \bibinfo{pages}{2278--2324}.
\newblock


\bibitem[\protect\citeauthoryear{Lee, Delbruck, and Pfeiffer}{Lee
  et~al\mbox{.}}{2016}]%
        {49}
\bibfield{author}{\bibinfo{person}{Jun~Haeng Lee}, \bibinfo{person}{Tobi
  Delbruck}, {and} \bibinfo{person}{Michael Pfeiffer}.}
  \bibinfo{year}{2016}\natexlab{}.
\newblock \showarticletitle{Training deep spiking neural networks using
  backpropagation}.
\newblock \bibinfo{journal}{\emph{Frontiers in Neuroscience}}
  \bibinfo{volume}{10} (\bibinfo{year}{2016}).
\newblock


\bibitem[\protect\citeauthoryear{Li, Shan, Hu, Wang, Chen, and Yang}{Li
  et~al\mbox{.}}{2013a}]%
        {3}
\bibfield{author}{\bibinfo{person}{Boxun Li}, \bibinfo{person}{Yi Shan},
  \bibinfo{person}{Miao Hu}, \bibinfo{person}{Yu Wang}, \bibinfo{person}{Yiran
  Chen}, {and} \bibinfo{person}{Huazhong Yang}.}
  \bibinfo{year}{2013}\natexlab{a}.
\newblock \showarticletitle{Memristor-based approximated computation}. In
  \bibinfo{booktitle}{\emph{Proceedings of the 2013 International Symposium on
  Low Power Electronics and Design}}. IEEE Press, \bibinfo{pages}{242--247}.
\newblock


\bibitem[\protect\citeauthoryear{Li, Shan, Hu, Wang, Chen, and Yang}{Li
  et~al\mbox{.}}{2013b}]%
        {li2013memristor-based}
\bibfield{author}{\bibinfo{person}{Boxun Li}, \bibinfo{person}{Yi Shan},
  \bibinfo{person}{Miao Hu}, \bibinfo{person}{Yu Wang}, \bibinfo{person}{Yiran
  Chen}, {and} \bibinfo{person}{Huazhong Yang}.}
  \bibinfo{year}{2013}\natexlab{b}.
\newblock \showarticletitle{Memristor-based approximated computation}. In
  \bibinfo{booktitle}{\emph{international symposium on low power electronics
  and design}}. \bibinfo{pages}{242--247}.
\newblock


\bibitem[\protect\citeauthoryear{LiKamWa, Hou, Gao, Polansky, and
  Zhong}{LiKamWa et~al\mbox{.}}{2016}]%
        {28}
\bibfield{author}{\bibinfo{person}{Robert LiKamWa}, \bibinfo{person}{Yunhui
  Hou}, \bibinfo{person}{Julian Gao}, \bibinfo{person}{Mia Polansky}, {and}
  \bibinfo{person}{Lin Zhong}.} \bibinfo{year}{2016}\natexlab{}.
\newblock \showarticletitle{RedEye: analog ConvNet image sensor architecture
  for continuous mobile vision}. In \bibinfo{booktitle}{\emph{Proceedings of
  the 43rd International Symposium on Computer Architecture}}. IEEE Press,
  \bibinfo{pages}{255--266}.
\newblock


\bibitem[\protect\citeauthoryear{Liu, Chen, Liu, Zhou, Zhou, Teman, Feng, Zhou,
  and Chen}{Liu et~al\mbox{.}}{2015a}]%
        {14}
\bibfield{author}{\bibinfo{person}{Daofu Liu}, \bibinfo{person}{Tianshi Chen},
  \bibinfo{person}{Shaoli Liu}, \bibinfo{person}{Jinhong Zhou},
  \bibinfo{person}{Shengyuan Zhou}, \bibinfo{person}{Olivier Teman},
  \bibinfo{person}{Xiaobing Feng}, \bibinfo{person}{Xuehai Zhou}, {and}
  \bibinfo{person}{Yunji Chen}.} \bibinfo{year}{2015}\natexlab{a}.
\newblock \showarticletitle{Pudiannao: A polyvalent machine learning
  accelerator}. In \bibinfo{booktitle}{\emph{ACM SIGARCH Computer Architecture
  News}}, Vol.~\bibinfo{volume}{43}. ACM, \bibinfo{pages}{369--381}.
\newblock


\bibitem[\protect\citeauthoryear{Liu, Du, Tao, Han, Luo, Xie, Chen, and
  Chen}{Liu et~al\mbox{.}}{2016}]%
        {liu2016cambricon}
\bibfield{author}{\bibinfo{person}{Shaoli Liu}, \bibinfo{person}{Zidong Du},
  \bibinfo{person}{Jinhua Tao}, \bibinfo{person}{Dong Han},
  \bibinfo{person}{Tao Luo}, \bibinfo{person}{Yuan Xie}, \bibinfo{person}{Yunji
  Chen}, {and} \bibinfo{person}{Tianshi Chen}.}
  \bibinfo{year}{2016}\natexlab{}.
\newblock \showarticletitle{Cambricon: An instruction set architecture for
  neural networks}. In \bibinfo{booktitle}{\emph{Proceedings of the 43rd
  International Symposium on Computer Architecture}}. IEEE Press,
  \bibinfo{pages}{393--405}.
\newblock


\bibitem[\protect\citeauthoryear{Liu, Mao, Liu, Li, Chen, Li, Wang, Jiang,
  Barnell, Wu, and Yang}{Liu et~al\mbox{.}}{2015b}]%
        {liu2015reno}
\bibfield{author}{\bibinfo{person}{Xiaoxiao Liu}, \bibinfo{person}{Mengjie
  Mao}, \bibinfo{person}{Beiye Liu}, \bibinfo{person}{Hai Li},
  \bibinfo{person}{Yiran Chen}, \bibinfo{person}{Boxun Li}, \bibinfo{person}{Yu
  Wang}, \bibinfo{person}{Hao Jiang}, \bibinfo{person}{Mark Barnell},
  \bibinfo{person}{Qing Wu}, {and} \bibinfo{person}{Jianhua Yang}.}
  \bibinfo{year}{2015}\natexlab{b}.
\newblock \showarticletitle{RENO: A high-efficient reconfigurable neuromorphic
  computing accelerator design}. In \bibinfo{booktitle}{\emph{Design Automation
  Conference (DAC), 2015 52nd ACM/EDAC/IEEE}}. IEEE, \bibinfo{pages}{1--6}.
\newblock


\bibitem[\protect\citeauthoryear{Mariet and Sra}{Mariet and Sra}{2015}]%
        {45}
\bibfield{author}{\bibinfo{person}{Zelda Mariet} {and} \bibinfo{person}{Suvrit
  Sra}.} \bibinfo{year}{2015}\natexlab{}.
\newblock \showarticletitle{Diversity Networks}.
\newblock \bibinfo{journal}{\emph{CoRR}}  \bibinfo{volume}{abs/1511.05077}
  (\bibinfo{year}{2015}).
\newblock
\urldef\tempurl%
\url{http://arxiv.org/abs/1511.05077}
\showURL{%
\tempurl}


\bibitem[\protect\citeauthoryear{Meier}{Meier}{2015}]%
        {43}
\bibfield{author}{\bibinfo{person}{Karlheinz Meier}.}
  \bibinfo{year}{2015}\natexlab{}.
\newblock \showarticletitle{A mixed-signal universal neuromorphic computing
  system}. In \bibinfo{booktitle}{\emph{Electron Devices Meeting (IEDM), 2015
  IEEE International}}. IEEE, \bibinfo{pages}{4--6}.
\newblock


\bibitem[\protect\citeauthoryear{Merolla, Arthur, Alvarez-Icaza, Cassidy,
  Sawada, Akopyan, Jackson, Imam, Guo, Nakamura, Brezzo, Vo, Esser, Appuswamy,
  Taba, Amir, Flickner, Risk, Manohar, and Modha}{Merolla
  et~al\mbox{.}}{2014}]%
        {8}
\bibfield{author}{\bibinfo{person}{Paul~A Merolla}, \bibinfo{person}{John~V
  Arthur}, \bibinfo{person}{Rodrigo Alvarez-Icaza}, \bibinfo{person}{Andrew~S
  Cassidy}, \bibinfo{person}{Jun Sawada}, \bibinfo{person}{Filipp Akopyan},
  \bibinfo{person}{Bryan~L Jackson}, \bibinfo{person}{Nabil Imam},
  \bibinfo{person}{Chen Guo}, \bibinfo{person}{Yutaka Nakamura},
  \bibinfo{person}{Bernard Brezzo}, \bibinfo{person}{lvan Vo},
  \bibinfo{person}{Steven~K Esser}, \bibinfo{person}{Rathinakumar Appuswamy},
  \bibinfo{person}{Brian Taba}, \bibinfo{person}{Arnon Amir},
  \bibinfo{person}{Myron~D Flickner}, \bibinfo{person}{William~P Risk},
  \bibinfo{person}{Rajit Manohar}, {and} \bibinfo{person}{Dharmendra~S Modha}.}
  \bibinfo{year}{2014}\natexlab{}.
\newblock \showarticletitle{A million spiking-neuron integrated circuit with a
  scalable communication network and interface}.
\newblock \bibinfo{journal}{\emph{Science}} \bibinfo{volume}{345},
  \bibinfo{number}{6197} (\bibinfo{year}{2014}), \bibinfo{pages}{668--673}.
\newblock


\bibitem[\protect\citeauthoryear{Mishkin and Matas}{Mishkin and Matas}{2015}]%
        {mishkin2015all}
\bibfield{author}{\bibinfo{person}{Dmytro Mishkin} {and} \bibinfo{person}{Jiri
  Matas}.} \bibinfo{year}{2015}\natexlab{}.
\newblock \showarticletitle{All you need is a good init}.
\newblock \bibinfo{journal}{\emph{arXiv preprint arXiv:1511.06422}}
  (\bibinfo{year}{2015}).
\newblock


\bibitem[\protect\citeauthoryear{Nair and Hinton}{Nair and Hinton}{2010}]%
        {nair2010rectified}
\bibfield{author}{\bibinfo{person}{Vinod Nair} {and}
  \bibinfo{person}{Geoffrey~E Hinton}.} \bibinfo{year}{2010}\natexlab{}.
\newblock \showarticletitle{Rectified linear units improve restricted boltzmann
  machines}. In \bibinfo{booktitle}{\emph{Proceedings of the 27th international
  conference on machine learning (ICML-10)}}. \bibinfo{pages}{807--814}.
\newblock


\bibitem[\protect\citeauthoryear{Prezioso, Merrikh-Bayat, Hoskins, Adam,
  Likharev, and Strukov}{Prezioso et~al\mbox{.}}{2015}]%
        {1}
\bibfield{author}{\bibinfo{person}{Mirko Prezioso}, \bibinfo{person}{Farnood
  Merrikh-Bayat}, \bibinfo{person}{BD Hoskins}, \bibinfo{person}{GC Adam},
  \bibinfo{person}{Konstantin~K Likharev}, {and} \bibinfo{person}{Dmitri~B
  Strukov}.} \bibinfo{year}{2015}\natexlab{}.
\newblock \showarticletitle{Training and operation of an integrated
  neuromorphic network based on metal-oxide memristors}.
\newblock \bibinfo{journal}{\emph{Nature}} \bibinfo{volume}{521},
  \bibinfo{number}{7550} (\bibinfo{year}{2015}), \bibinfo{pages}{61--64}.
\newblock


\bibitem[\protect\citeauthoryear{Qadeer, Hameed, Shacham, Venkatesan,
  Kozyrakis, and Horowitz}{Qadeer et~al\mbox{.}}{2013}]%
        {26}
\bibfield{author}{\bibinfo{person}{Wajahat Qadeer}, \bibinfo{person}{Rehan
  Hameed}, \bibinfo{person}{Ofer Shacham}, \bibinfo{person}{Preethi
  Venkatesan}, \bibinfo{person}{Christos Kozyrakis}, {and}
  \bibinfo{person}{Mark~A Horowitz}.} \bibinfo{year}{2013}\natexlab{}.
\newblock \showarticletitle{Convolution engine: balancing efficiency \&
  flexibility in specialized computing}. In \bibinfo{booktitle}{\emph{ACM
  SIGARCH Computer Architecture News}}, Vol.~\bibinfo{volume}{41}. ACM,
  \bibinfo{pages}{24--35}.
\newblock


\bibitem[\protect\citeauthoryear{Reagen, Whatmough, Adolf, Rama, Lee, Lee,
  Hern{\'a}ndez-Lobato, Wei, and Brooks}{Reagen et~al\mbox{.}}{2016}]%
        {27}
\bibfield{author}{\bibinfo{person}{Brandon Reagen}, \bibinfo{person}{Paul
  Whatmough}, \bibinfo{person}{Robert Adolf}, \bibinfo{person}{Saketh Rama},
  \bibinfo{person}{Hyunkwang Lee}, \bibinfo{person}{Sae~Kyu Lee},
  \bibinfo{person}{Jos{\'e}~Miguel Hern{\'a}ndez-Lobato},
  \bibinfo{person}{Gu-Yeon Wei}, {and} \bibinfo{person}{David Brooks}.}
  \bibinfo{year}{2016}\natexlab{}.
\newblock \showarticletitle{Minerva: Enabling low-power, highly-accurate deep
  neural network accelerators}. In \bibinfo{booktitle}{\emph{Proceedings of the
  43rd International Symposium on Computer Architecture}}. IEEE Press,
  \bibinfo{pages}{267--278}.
\newblock


\bibitem[\protect\citeauthoryear{Shafiee, Nag, Muralimanohar, Balasubramonian,
  Strachan, Hu, Williams, and Srikumar}{Shafiee et~al\mbox{.}}{2016}]%
        {18}
\bibfield{author}{\bibinfo{person}{Ali Shafiee}, \bibinfo{person}{Anirban Nag},
  \bibinfo{person}{Naveen Muralimanohar}, \bibinfo{person}{Rajeev
  Balasubramonian}, \bibinfo{person}{John~Paul Strachan}, \bibinfo{person}{Miao
  Hu}, \bibinfo{person}{R~Stanley Williams}, {and} \bibinfo{person}{Vivek
  Srikumar}.} \bibinfo{year}{2016}\natexlab{}.
\newblock \showarticletitle{ISAAC: A convolutional neural network accelerator
  with in-situ analog arithmetic in crossbars}. In
  \bibinfo{booktitle}{\emph{Proceedings of the 43rd International Symposium on
  Computer Architecture}}. IEEE Press, \bibinfo{pages}{14--26}.
\newblock


\bibitem[\protect\citeauthoryear{Sharma, Park, Mahajan, Amaro, Kim, Shao,
  Mishra, and Esmaeilzadeh}{Sharma et~al\mbox{.}}{2016}]%
        {sharma2016high}
\bibfield{author}{\bibinfo{person}{Hardik Sharma}, \bibinfo{person}{Jongse
  Park}, \bibinfo{person}{Divya Mahajan}, \bibinfo{person}{Emmanuel Amaro},
  \bibinfo{person}{Joon~Kyung Kim}, \bibinfo{person}{Chenkai Shao},
  \bibinfo{person}{Asit Mishra}, {and} \bibinfo{person}{Hadi Esmaeilzadeh}.}
  \bibinfo{year}{2016}\natexlab{}.
\newblock \showarticletitle{From high-level deep neural models to FPGAs}. In
  \bibinfo{booktitle}{\emph{Microarchitecture (MICRO), 2016 49th Annual
  IEEE/ACM International Symposium on}}. IEEE, \bibinfo{pages}{1--12}.
\newblock


\bibitem[\protect\citeauthoryear{Shi, Pei, Deng, Wang, Deng, Wang, Zhang, Chen,
  Zhao, Song, Zeng, Li, Li, and Ma}{Shi et~al\mbox{.}}{2015}]%
        {11}
\bibfield{author}{\bibinfo{person}{L. Shi}, \bibinfo{person}{J. Pei},
  \bibinfo{person}{N. Deng}, \bibinfo{person}{D. Wang}, \bibinfo{person}{L.
  Deng}, \bibinfo{person}{Y. Wang}, \bibinfo{person}{Y. Zhang},
  \bibinfo{person}{F. Chen}, \bibinfo{person}{M. Zhao}, \bibinfo{person}{S.
  Song}, \bibinfo{person}{F. Zeng}, \bibinfo{person}{G. Li},
  \bibinfo{person}{H. Li}, {and} \bibinfo{person}{C. Ma}.}
  \bibinfo{year}{2015}\natexlab{}.
\newblock \showarticletitle{Development of a neuromorphic computing system}. In
  \bibinfo{booktitle}{\emph{2015 IEEE International Electron Devices Meeting
  (IEDM)}}. \bibinfo{pages}{4.3.1--4.3.4}.
\newblock
\urldef\tempurl%
\url{https://doi.org/10.1109/IEDM.2015.7409624}
\showDOI{\tempurl}


\bibitem[\protect\citeauthoryear{Shin, Lee, Lee, and Yoo}{Shin
  et~al\mbox{.}}{2017}]%
        {DNPU2017}
\bibfield{author}{\bibinfo{person}{Dongjoo Shin}, \bibinfo{person}{Jinmook
  Lee}, \bibinfo{person}{Jinsu Lee}, {and} \bibinfo{person}{HoiJun Yoo}.}
  \bibinfo{year}{2017}\natexlab{}.
\newblock \showarticletitle{DNPU: An 8.1TOPS/W Reconfigurable CNN-RNN Processor
  for General Purpose Deep Neural Networks}. In
  \bibinfo{booktitle}{\emph{International Solid-State Circuits Conference}}.
  IEEE.
\newblock


\bibitem[\protect\citeauthoryear{Simonyan and Zisserman}{Simonyan and
  Zisserman}{2014}]%
        {VGG16}
\bibfield{author}{\bibinfo{person}{Karen Simonyan} {and}
  \bibinfo{person}{Andrew Zisserman}.} \bibinfo{year}{2014}\natexlab{}.
\newblock \showarticletitle{Very deep convolutional networks for large-scale
  image recognition}.
\newblock \bibinfo{journal}{\emph{arXiv preprint arXiv:1409.1556}}
  (\bibinfo{year}{2014}).
\newblock


\bibitem[\protect\citeauthoryear{Song, Qian, Li, and Chen}{Song
  et~al\mbox{.}}{2016a}]%
        {songpipelayer}
\bibfield{author}{\bibinfo{person}{Linghao Song}, \bibinfo{person}{Xuehai
  Qian}, \bibinfo{person}{Hai Li}, {and} \bibinfo{person}{Yiran Chen}.}
  \bibinfo{year}{2016}\natexlab{a}.
\newblock \showarticletitle{PipeLayer: A Pipelined ReRAM-Based Accelerator for
  Deep Learning}. In \bibinfo{booktitle}{\emph{High Performance Computer
  Architecture (HPCA),2017 23rd IEEE Symposium on}}. IEEE.
\newblock


\bibitem[\protect\citeauthoryear{Song, Wang, Han, Zhao, Liu, and Li}{Song
  et~al\mbox{.}}{2016b}]%
        {25}
\bibfield{author}{\bibinfo{person}{Lili Song}, \bibinfo{person}{Ying Wang},
  \bibinfo{person}{Yinhe Han}, \bibinfo{person}{Xin Zhao},
  \bibinfo{person}{Bosheng Liu}, {and} \bibinfo{person}{Xiaowei Li}.}
  \bibinfo{year}{2016}\natexlab{b}.
\newblock \showarticletitle{C-Brain: A deep learning accelerator that tames the
  diversity of CNNs through adaptive data-level parallelization}. In
  \bibinfo{booktitle}{\emph{Design Automation Conference (DAC), 2016 53nd
  ACM/EDAC/IEEE}}. IEEE, \bibinfo{pages}{1--6}.
\newblock


\bibitem[\protect\citeauthoryear{Vanhoucke, Senior, and Mao}{Vanhoucke
  et~al\mbox{.}}{2011}]%
        {vanhoucke2011improving}
\bibfield{author}{\bibinfo{person}{Vincent Vanhoucke}, \bibinfo{person}{Andrew
  Senior}, {and} \bibinfo{person}{Mark~Z Mao}.}
  \bibinfo{year}{2011}\natexlab{}.
\newblock \showarticletitle{Improving the speed of neural networks on CPUs}. In
  \bibinfo{booktitle}{\emph{Proc. Deep Learning and Unsupervised Feature
  Learning NIPS Workshop}}, Vol.~\bibinfo{volume}{1}. Citeseer,
  \bibinfo{pages}{4}.
\newblock


\bibitem[\protect\citeauthoryear{Wendt, Ehrlich, and Sch\"{u}ffny}{Wendt
  et~al\mbox{.}}{2008}]%
        {42}
\bibfield{author}{\bibinfo{person}{Karsten Wendt}, \bibinfo{person}{Matthias
  Ehrlich}, {and} \bibinfo{person}{Ren{\'e} Sch\"{u}ffny}.}
  \bibinfo{year}{2008}\natexlab{}.
\newblock \showarticletitle{A Graph Theoretical Approach for a Multistep
  Mapping Software for the FACETS Project}. In
  \bibinfo{booktitle}{\emph{Proceedings of the 2Nd WSEAS International
  Conference on Computer Engineering and Applications}}
  \emph{(\bibinfo{series}{CEA'08})}. \bibinfo{publisher}{World Scientific and
  Engineering Academy and Society (WSEAS)}, \bibinfo{address}{Stevens Point,
  Wisconsin, USA}, \bibinfo{pages}{189--194}.
\newblock
\showISBNx{978-960-6766-33-6}
\urldef\tempurl%
\url{http://dl.acm.org/citation.cfm?id=1373936.1373969}
\showURL{%
\tempurl}


\bibitem[\protect\citeauthoryear{Zhang, Li, Sun, Guan, Xiao, and Cong}{Zhang
  et~al\mbox{.}}{2015}]%
        {24}
\bibfield{author}{\bibinfo{person}{Chen Zhang}, \bibinfo{person}{Peng Li},
  \bibinfo{person}{Guangyu Sun}, \bibinfo{person}{Yijin Guan},
  \bibinfo{person}{Bingjun Xiao}, {and} \bibinfo{person}{Jason Cong}.}
  \bibinfo{year}{2015}\natexlab{}.
\newblock \showarticletitle{Optimizing fpga-based accelerator design for deep
  convolutional neural networks}. In \bibinfo{booktitle}{\emph{Proceedings of
  the 2015 ACM/SIGDA International Symposium on Field-Programmable Gate
  Arrays}}. ACM, \bibinfo{pages}{161--170}.
\newblock


\bibitem[\protect\citeauthoryear{Zhang, Du, Zhang, Lan, Liu, Li, Guo, Chen, and
  Chen}{Zhang et~al\mbox{.}}{2016}]%
        {zhang2016cambricon}
\bibfield{author}{\bibinfo{person}{Shijin Zhang}, \bibinfo{person}{Zidong Du},
  \bibinfo{person}{Lei Zhang}, \bibinfo{person}{Huiying Lan},
  \bibinfo{person}{Shaoli Liu}, \bibinfo{person}{Ling Li}, \bibinfo{person}{Qi
  Guo}, \bibinfo{person}{Tianshi Chen}, {and} \bibinfo{person}{Yunji Chen}.}
  \bibinfo{year}{2016}\natexlab{}.
\newblock \showarticletitle{Cambricon-X: An accelerator for sparse neural
  networks}. In \bibinfo{booktitle}{\emph{Microarchitecture (MICRO), 2016 49th
  Annual IEEE/ACM International Symposium on}}. IEEE, \bibinfo{pages}{1--12}.
\newblock


\end{thebibliography}



\end{document}